%% file: acl_latex.tex
\pdfoutput=1
\documentclass[11pt]{article}

\usepackage[final]{acl}
\usepackage{afterpage}
\usepackage{times}
\usepackage{latexsym}
\usepackage{listings}
\usepackage{graphicx}
\usepackage{subcaption}
\usepackage[T1]{fontenc}
\usepackage[utf8]{inputenc}
\usepackage{microtype}
\usepackage{inconsolata}
\usepackage{graphicx}
\usepackage{booktabs}  

\input{preamble}

\title{Auditing Chinese Web-scale Corpora via Sampled BPE Token Statistics\\
\small
\usepalatino{\textcolor{orange}{\warning{}Caution: this paper may include offensive and upsetting content.}}}

\author{
    \centerline{Qingjie Zhang\textsuperscript{1}, \ 
    Ziqi Tang\textsuperscript{1}, \ 
    Jie Zhang\textsuperscript{2}, \ 
    Gelei Deng\textsuperscript{3}, \
    Jinfeng Li\textsuperscript{4}, 
    } \vspace{0.5mm}  \\
    \centerline{\textbf{
    Yuefeng Chen\textsuperscript{4}, \ 
    Yitong Yang\textsuperscript{4}, \ 
    Hui Xue\textsuperscript{4}, \
    Tianwei Zhang\textsuperscript{3}, \ 
    and Han Qiu\textsuperscript{1*}
    }} \vspace{0.5mm} \\
    \centerline{\normalsize{$^{1}$Tsinghua University~~$^{2}$SiliconProspect AI~~$^{3}$Nanyang Technological University~~$^{4}$Alibaba Group}} \vspace{0.5mm} \\
    \centerline{\normalsize{Emails: \{qj-zhang24@mails., qiuhan@\}tsinghua.edu.cn~~\textsuperscript{*}Corresponding author}}
}

\begin{document}
\begin{CJK}{UTF8}{gbsn} 
\maketitle
\begin{abstract}
Chinese web pollution has surfaced in LLMs, motivating audits of upstream Chinese corpora.
However, auditing such corpora faces three challenges: (1) their web-scale size makes full scan costly; (2) prior analyses are often too coarse to expose token-level pollution; (3) Chinese web pollution is implicit and rapidly changing.
We propose \textsc{Sampled-BPE}, a lightweight token-level auditing pipeline that samples a small subset and trains BPE tokenizer to surface polluted tokens. 
Experiments show that \textsc{Sampled-BPE} preserves usable estimates while substantially reducing runtime and memory: a 148.4$\times$ speedup and a 35.8$\times$ memory reduction induce only 4.25\% relative error for pollution categories.
We apply the pipeline to 11 open Chinese corpora and 6 Chinese Common Crawl snapshots from 2021 to 2026. The audit reveals widespread but uneven pollution across open corpora, as well as highly polluted and temporally shifting Chinese web content. 
We further release a hierarchical Chinese web token dataset\footnote{\url{https://github.com/qingjiesjtu/SampledBPE}} with 630k+ token records, each with web context, category, and explanation fields, organized as 92k+ trees to support review and tracing of pollution.
\end{abstract}

\section{Introduction}
Chinese web pollution has surfaced in LLMs, including spam- and pornography-related Chinese tokens in ChatGPT's vocabulary~\citep{yang2024gpt4oChineseTokenPollution} and Chinese gambling content in Codex outputs~\citep{openaiCommunity2026codexChineseGambling}. Together with recent work linking polluted Chinese tokens to corpus pollution~\citep{zhang2025speculating}, these observations motivate auditing upstream Chinese corpora to understand where such pollution originates, how prevalent it is, and how it may propagate into downstream datasets.

However, auditing upstream Chinese corpora remains difficult. \emph{First}, mainstream Chinese corpora are often at web scale, consisting of hundreds of gigabytes to terabytes of text \citep{yuan2021wudaocorpora,chen2023chinesewebtext,he2023wanjuan,oepen2025hplt3}. Fully scanning such corpora is costly \citep{penedo2023refinedweb,penedo2024fineweb,abbas2023semdedup}, e.g., 99 days for indexing 83TB data on a 128-core CPU node \citep{xu-etal-2025-infini}. \emph{Second}, existing corpus analyses often remain coarse-grained such as language distribution, source domains, or quality scores at document-level \citep{dodge2021documentingc4,kreutzer2022quality,soldaini2024dolma,hagar2025practical}. These views describe broad corpus composition, but cannot reveal fine-grained pollution at token-level. \emph{Third}, Chinese web pollution is often implicit and rapidly changing \citep{macavaney2019hate,jiang2021neologisms,zhang2025speculating}. A predefined keyword list can miss newly emerging polluted tokens and quickly become stale. Together, these challenges call for \textbf{\textit{a lightweight, token-level, periodic auditing pipeline}}.

\begin{figure}[t]
    \centering
    \includegraphics[width=0.9\linewidth]{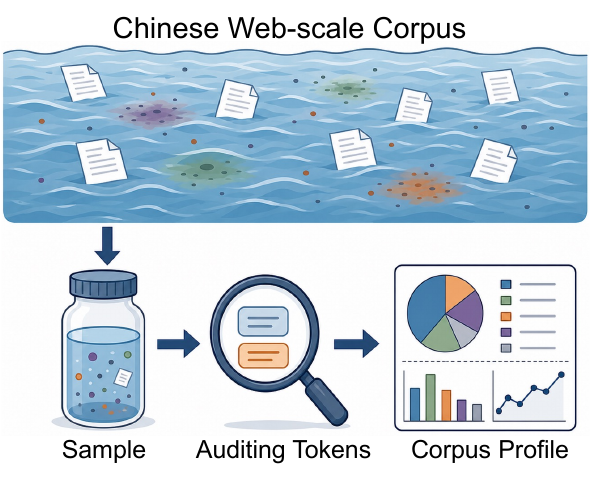}
    \vspace{-1em}
    \caption{Intuition from marine pollution monitoring: sample a subset, train a tokenizer, map token-level pollution signals, and estimate a corpus profile.}
    \label{fig:auditing_idea}
    \vspace{-2em}
\end{figure}

Inspired by marine pollution monitoring (illustrated in \autoref{fig:auditing_idea}), which estimates overall pollution from sampled observations \citep{karydis2013marinewater}, we propose \textsc{Sampled-BPE}, a simple yet effective pipeline for auditing web-scale Chinese corpora when full inspection is too costly. Specifically, we sample a small subset, train a byte-pair encoding (BPE) tokenizer \citep{sennrich2016neural} to surface and count high-frequency tokens, map these tokens to content categories using Internet search evidence, and aggregate token statistics into corpus-level profiles. This design avoids relying on a fixed keyword list, keeps the audit at token level, and makes repeated auditing feasible as Chinese web pollution evolves.

Experiments show that \textsc{Sampled-BPE} preserves usable token-level and category-level estimates while substantially reducing runtime and memory. Using this pipeline, we find that pollution is widespread but uneven across open Chinese corpora. We further show that Chinese Common Crawl \citep{commoncrawl} is highly polluted and shifts over time, indicating the need for periodic upstream auditing. Finally, we release a hierarchical Chinese web token dataset to make these token-level findings reviewable. 

This work makes three contributions:
\begin{packeditemize}
    \item \textbf{Method.} We propose and validate \textsc{Sampled-BPE}, a lightweight pipeline for token-level auditing of web-scale Chinese corpora. \textit{It reduces token-level auditing cost from months to hours while preserving usable estimates}.
    \item \textbf{Results.} We systematically audit \textit{11 open Chinese corpora and 6 upstream Chinese Common Crawl snapshots from 2021 to 2026}, revealing widespread and shifting pollution. For example, 68.72\% of 2026 Chinese Common Crawl snapshot is Adult Content.
    \item \textbf{Dataset.} We release a \textit{hierarchical Chinese web token dataset} containing 630{,}684 token records, each with web context, category, and explanation fields. The dataset organizes tokens into 92{,}972 trees, supporting review and tracing of Internet pollution.
\end{packeditemize}

\section{Background}

In this section, we first summarize the corpora audited in this paper, then discuss related work on corpus auditing and Chinese web pollution.

\subsection{Open Chinese Corpora and Landscape}
\label{sec:openChineseCorpora}

\begin{figure}[t]
    \centering
    \includegraphics[width=0.95\linewidth]{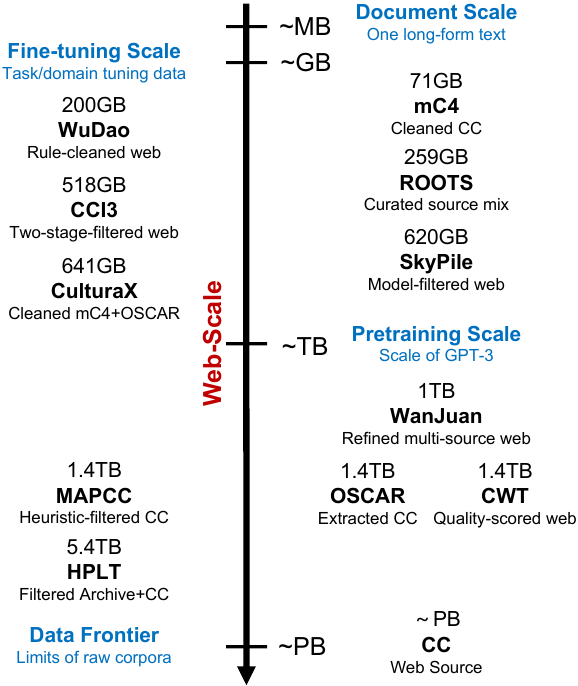}
    \caption{Scale landscape of Chinese web corpora audited in this work.}
    \label{fig:corpus}
    \vspace{-3ex}
\end{figure}

LLM corpora span from MB-scale documents, to GB-scale tuning data, TB-scale pretraining corpora, and PB-scale upstream web archives~\citep{wang2018glue,longpre2023flan,brown2020language,commoncrawl}. In this work, we use \textit{web-scale} to refer to corpora in the TB range \citep{li2025ticlm,roziewski2016languagecrawl}. \autoref{fig:corpus} places the Chinese corpora audited in this work within this scale landscape.

Mainstream open Chinese corpora include Chinese portions of broad multilingual corpora, such as OSCAR~\citep{abadji2022oscar}, mC4~\citep{xue2021mt5}, HPLT~\citep{oepen2025hplt3}, CulturaX~\citep{nguy2024culturax}, and ROOTS~\citep{laurencon2022roots}; and Chinese-specific releases, such as CWT~\citep{chen2023chinesewebtext}, WanJuan~\citep{he2023wanjuan}, MAPCC~\citep{du2024chinesetinyllm}, SkyPile~\citep{wei2023skywork}, CCI3~\citep{wang2024cci30hq}, and WuDao~\citep{yuan2021wudaocorpora}. Detailed descriptions are provided in \autoref{app:chinese-corpora}.

\subsection{Related Work}

\begin{figure*}[t]
    \centering
    \includegraphics[width=0.95\textwidth]{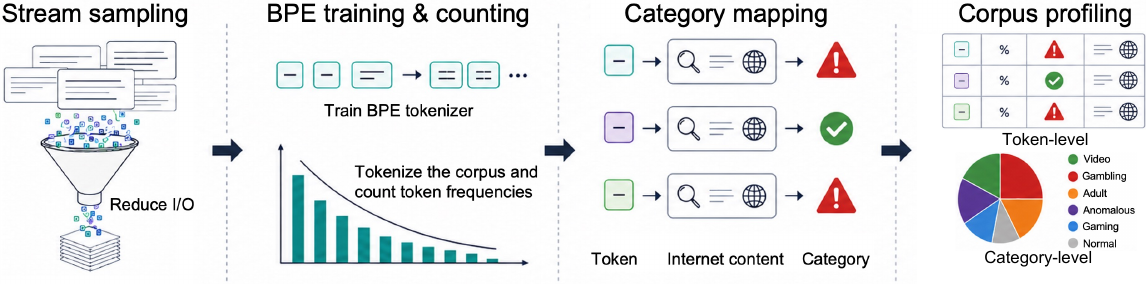}
    \caption{Conceptual overview of the \textsc{Sampled-BPE} auditing pipeline. It samples web-scale corpora, trains and counts BPE tokens, maps tokens to content categories with Internet evidence, and aggregates token statistics into corpus pollution profiles.}
    \label{fig:auditing_pipeline}
    \vspace{-3ex}
\end{figure*}

\paragraph{Corpus Curation and Auditing.}
Web-scale corpora are usually curated from web text through source selection, language identification, quality filtering, safety or blocklist filtering, and deduplication~\citep{penedo2024fineweb,soldaini2024dolma,weber2024redpajama,penedo2023refinedweb}. Prior documentation and auditing work further improves transparency and reveals residual risks such as unexpected sources, low-quality text, duplication, contamination, and filtering side effects~\citep{lee2022deduplicating,kreutzer2022quality,dodge2021documentingc4,gebru2021datasheets}. These studies mainly examine source-level, document-level, quality-level, or deduplication-level properties. In contrast, \textit{our work audits corpora at token-level}.

\paragraph{Chinese Web Pollution and Polluted Tokens.}
The research community has reported multiple cases where Chinese web pollution surfaces in LLMs. For example, it appears in GPT vocabularies: after GPT-4o was released, researchers reported that many of its longest Chinese tokens were not ordinary words, but long phrases associated with spam, pornography, gambling, or scams~\citep{yang2024gpt4oChineseTokenPollution}. Moreover, it can even appear in outputs for otherwise normal tasks: users reported Codex outputs that unexpectedly inserted Chinese gambling or lottery-like strings during coding contexts, including around \texttt{apply\_patch} outputs and malformed tool-call text~\citep{openaiCommunity2026codexChineseGambling}.
Recent work studies this issue through polluted tokens in LLM vocabularies, connecting their presence to possible training corpora pollution~\citep{zhang2025speculating}.
However, these observations inspect downstream models rather than upstream corpora.

\section{Sampled-BPE}

To enable low-cost and accurate pollution auditing of web-scale Chinese corpora, \textsc{Sampled-BPE} does not aim to reconstruct the full vocabulary. Instead, it approximates the token-level and category-level statistics of corpora.

\subsection{Auditing Pipeline}
\label{sec:auditing-pipeline}

\autoref{fig:auditing_pipeline} gives a conceptual overview of the four-stage auditing pipeline.

\begin{figure*}[t]
    \centering
    \includegraphics[width=\textwidth]{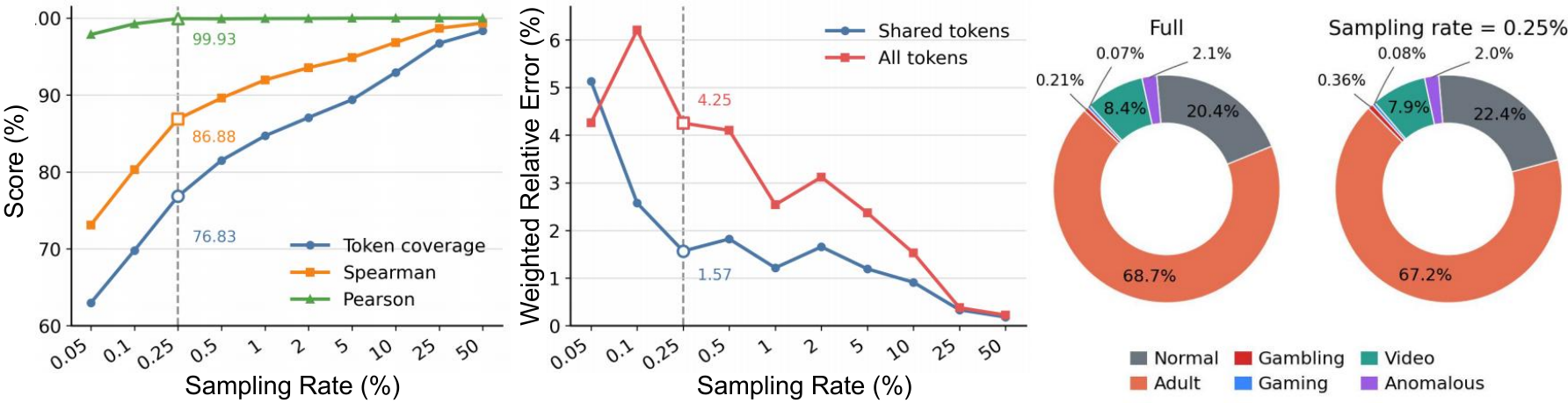}
    \caption{Sampling preserves auditing accuracy. \textbf{Left}: token-level coverage and token-ratio correlations, with the 0.25\% sampling rate annotated. \textbf{Middle}: category-level weighted relative error, with the 0.25\% sampling rate annotated. \textbf{Right}: category composition of 0.25\% sampling rate compared with \texttt{Full}.}
    \label{fig:sampling_accuracy}
    \vspace{-3ex}
\end{figure*}

\paragraph{Step 1: streaming sampling.} Given a target corpus, we sample documents in a single sequential pass. Compared with random sampling which requires enumerating or shuffling document identifiers, streaming sampling makes the sampling cost nearly linear in the input scan. This system choice reduces I/O and avoids full materialization of large corpus, while yielding audit accuracy comparable to random sampling (see detailed comparison in \autoref{app:streaming-vs-random}).

\paragraph{Step 2: BPE training \& counting.} We train a BPE tokenizer on the selected corpus and use the same tokenizer to count token frequencies \citep{sennrich2016neural}. We use BPE for three reasons. \emph{First}, because BPE is a compression method that repeatedly merges frequent adjacent units, its learned vocabulary surfaces recurrent lexical patterns that characterize the corpus. \emph{Second}, high-frequency polluted tokens can be exposed without relying on a fixed keyword list, allowing new pollution terms to be discovered \citep{macavaney2019hate,jiang2021neologisms,zhang2025speculating}. \emph{Third}, BPE is efficient for web-scale auditing: training is driven by local adjacent-pair frequency statistics under a fixed merge budget, and counting only requires a linear pass with the learned tokenizer rather than expensive document-level semantic analysis.

\paragraph{Step 3: category mapping.} Polluted tokens are often subtle, abbreviated, or context-dependent: the token alone may not reveal its category. Therefore, inspired by prior work on Chinese polluted-token detection \citep{zhang2025speculating}, we map tokens, with Internet search results as contextual evidence, into one of six content categories: \textit{Normal Content, Adult Content, Online Gambling, Online Gaming, Online Video, and Anomalous}. To automate this step, we use GLM-4-32B, an open-source Chinese LLM with strong Chinese comprehension ability, as the base category classifier. The model is fine-tuned to predict a category from each token and its corresponding Internet search results, using a train/test split constructed from expert annotations of GPT Chinese vocabularies in \citet{zhang2025speculating}. It achieves 97.32\% classification accuracy on the test split after fine-tuning. Search-evidence settings and category-mapping robustness analyses are provided in \autoref{app:category-robustness}.

\paragraph{Step 4: corpus profiling.} We combine token frequencies with category assignments to build a corpus profile. At the token level, the profile records each token's ratio, predicted category, Internet evidence, and classification rationale. This enables later analyses of shared polluted tokens, corpus-specific polluted terms, and high-frequency lexical evolution. At the category level, we aggregate token ratios into category prevalence and total pollution ratios, which are the quantities reported in later corpus-level comparisons and temporal analyses.

We refer to the unsampled corpus as \texttt{Full}. In the following experiments, each sampled corpus is audited with the same pipeline and compared against \texttt{Full} to evaluate whether sampling preserves the token-level and category-level statistics needed for pollution analysis.

\subsection{Sampling Preserves Auditing Accuracy}

We first ask whether \textsc{Sampled-BPE} preserves the token-level and category-level statistics needed to audit corpus pollution, that is, whether sampling gives an acceptable approximation to \texttt{Full}.

For each sampling rate, we compare the sampled corpus with \texttt{Full}. Let $V_{\mathrm{Full}}$ denote the token set extracted from \texttt{Full}, and let $V_{\mathrm{sample}}$ denote the token set extracted from a sampled corpus. At the token level, we measure token coverage as $|V_{\mathrm{Full}}\cap V_{\mathrm{sample}}|/|V_{\mathrm{Full}}|$, and compute Spearman and Pearson correlations between token ratios over shared tokens $V_{\mathrm{Full}}\cap V_{\mathrm{sample}}$. At the category level, after mapping tokens into content categories, we compute weighted relative error both over shared tokens and over all tokens.

\autoref{fig:sampling_accuracy} (Left) shows that sampled corpora still provide an acceptable token-level approximation to \texttt{Full} at low sampling rates. For example, at the 0.25\% sampling rate, token coverage remains: the sampled corpus recovers 76.83\% of \texttt{Full} tokens. More importantly, the ratios of the covered tokens closely match \texttt{Full}, with Spearman reaching 86.88\% and Pearson reaching 99.93\%. This suggests that \textit{\textbf{sampling preserves an acceptable approximation of token-level statistics}}.

\autoref{fig:sampling_accuracy} (Middle) shows that the same trend appears after aggregating tokens into content categories. Even when token coverage is imperfect, the weighted relative error remains around 5\% at the lowest sampling rate, for both shared tokens and all tokens. \autoref{fig:sampling_accuracy} (Right) further illustrates this at the 0.25\% sampling rate, where the category composition remains close to the \texttt{Full} distribution. This shows that \textit{\textbf{sampling also preserves category-level statistics for pollution auditing}}.

\subsection{Sampling Reduces Auditing Cost}

\begin{figure}[t]
    \centering
    \includegraphics[width=0.95\linewidth]{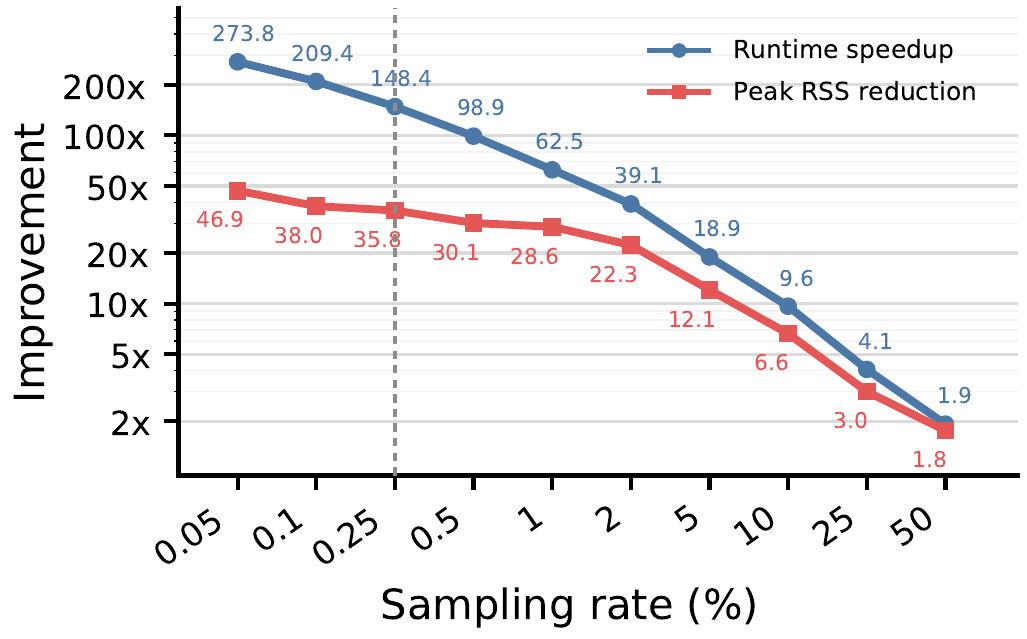}
    \vspace{-1ex}
    \caption{Runtime and memory improvement across sampling rates for \textsc{Sampled-BPE}.}
    \label{fig:sampling_cost}
    \vspace{-3ex}
\end{figure}

\begin{table*}[t]
\centering
\scriptsize
\resizebox{\linewidth}{!}{
\begin{tabular}{lrrrrrrrrrrr}
\toprule
 & OSCAR & mC4 & HPLT & CulturaX & CWT & ROOTS & WanJuan & MAPCC & SkyPile & CCI3 & WuDao \\
\midrule
\textbf{Pollution} & \textbf{83.38} & \textbf{25.41} & \textbf{4.95} & \textbf{3.36} & \textbf{2.35} & \textbf{2.02} & \textbf{0.75} & \textbf{0.69} & \textbf{0.61} & \textbf{0.55} & \textbf{0.50} \\
Adult & 76.10 & 0.31 & 1.95 & 0.15 & 0.05 & 0.23 & 0.04 & 0.10 & 0.06 & 0.16 & 0.10 \\
Gambling & 0.56 & 23.17 & 1.61 & 2.45 & 0.22 & 0.00 & 0.15 & 0.08 & 0.04 & 0.00 & 0.01 \\
Gaming & 0.16 & 0.46 & 0.25 & 0.16 & 0.29 & 0.00 & 0.17 & 0.09 & 0.05 & 0.01 & 0.06 \\
Video & 5.02 & 0.32 & 0.72 & 0.18 & 0.06 & 0.00 & 0.05 & 0.04 & 0.04 & 0.00 & 0.01 \\
Anomalous & 1.53 & 1.15 & 0.43 & 0.42 & 1.73 & 1.79 & 0.33 & 0.38 & 0.42 & 0.37 & 0.31 \\
\bottomrule
\end{tabular}
}
\caption{Pollution ratios (\%) for the Chinese portions of 11 representative open corpora, sorted by total pollution.}
\label{tab:zh_corpora_pollution}
\vspace{-2ex}
\end{table*}

Having shown that sampling preserves the statistics needed for pollution auditing, we next examine its runtime speedup and memory reduction.

We measure end-to-end auditing cost as BPE training time plus token-counting time, and we report peak RSS as the memory footprint. All runs are conducted on the same machine with 96 CPU cores and 768 GB RAM.

\autoref{fig:sampling_cost} shows that smaller samples sharply reduce both runtime and memory. At the 0.25\% sampling rate, the auditing pipeline achieves a 148.4$\times$ runtime speedup. The memory reduction is also substantial, with peak RSS reduced by 35.8$\times$. Extremely small samples provide even larger savings, but \autoref{fig:sampling_accuracy} shows that their token coverage and category estimates are less accurate. Larger samples improve accuracy, but their computational advantage decreases.

Overall, sampling makes web-scale corpus auditing \textit{\textbf{substantially cheaper while retaining acceptable accuracy for pollution analysis}}. The time cost of auditing 1TB corpus can be reduced from months to hours.

\section{Pollution in Open Chinese Corpora}
\label{sec:open-corpus-audits}

This section audits pollution in 11 open Chinese corpora or the Chinese portions of multilingual corpora as stated in \autoref{sec:openChineseCorpora}. Following \citep{zhang2025speculating}, percentages are computed over tokens containing at least three Chinese characters over six categories. Release provenance and upstream-processing details are provided in \autoref{app:chinese-corpora}.

\subsection{Pollution Is Widespread but Uneven}

\autoref{tab:zh_corpora_pollution} shows that each corpus contains polluted tokens, however, the pollution ratio varies sharply, ranging from 0.50\% in the cleanest corpora to 83.38\%. OSCAR is the highest-pollution corpus, and mC4 is also heavily polluted at 25.41\%. By contrast, WanJuan, MAPCC, SkyPile, CCI3, and WuDao all remain below 1\% total pollution. This shows that \textit{\textbf{Chinese pollution is widespread across open corpora}}, but is highly uneven across corpus families and construction pipelines.

The lower pollution ratios in several corpora suggest that cleaning and curation substantially reduce pollution. Broad multilingual web pipelines such as OSCAR, mC4, HPLT, and CulturaX have the highest pollution ratios in \autoref{tab:zh_corpora_pollution}. In contrast, corpora such as WanJuan, MAPCC, SkyPile, CCI3, and WuDao, which emphasize Chinese-specific collection, trusted sources, or quality filtering, are much cleaner. At the same time, even the lowest-pollution corpora retain measurable polluted tokens, indicating that \textit{\textbf{cleaning helps, but does not eliminate pollution}}. To support this claim, \autoref{app:chinese-cleaning} gives two more focused comparisons between cleaned and uncleaned corpus variants.

The category rows in \autoref{tab:zh_corpora_pollution} show that pollution is not monolithic. OSCAR is dominated by Adult Content, which accounts for 76.10\%. mC4 has a different profile: Online Gambling alone reaches 23.17\%. HPLT contains both Adult Content and Online Gambling at non-trivial levels. Cleaner corpora exhibit a different residual profile: their remaining pollution is usually not dominated by a single category; instead, the largest residual component is often Anomalous, reflecting rare, peculiar, or contextually irrelevant phrases that survive quality filtering. These contrasts show that \textit{\textbf{the composition of pollution varies across corpora}}. We provide word-cloud views to show these corpus differences in \autoref{app:wordclouds}.

\subsection{Pollution Is Shared, Yet Mostly Unique}
\label{sec:token_level_pollution}

Since open Chinese corpora differ sharply in both total pollution and category composition, we further investigate whether the same polluted tokens recur across corpora or whether each corpus has its own lexical artifacts. 

To compare polluted tokens across the 11 representative corpus in \autoref{tab:zh_corpora_pollution}, we define a directional coverage metric:
\[
\mathrm{Coverage}(A \rightarrow B) = \frac{|P_A \cap P_B|}{|P_A|}.
\]
Here $P_A$ and $P_B$ denote the polluted tokens in corpora $A$ and $B$.

\begin{figure}[t]
    \centering
    \includegraphics[width=\linewidth]{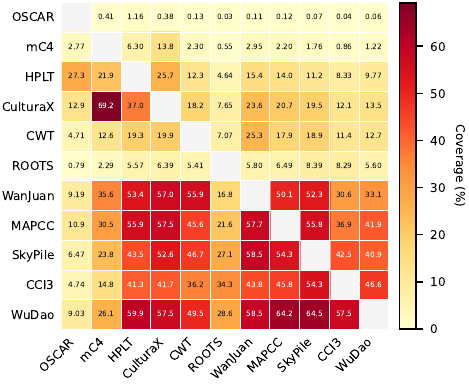}
    \caption{Pairwise coverage of polluted tokens across corpora. Each cell (omitting the diagonal) reports the percentage of polluted tokens in the row corpus that also appear in the column corpus.}
    \label{fig:polluted_token_coverage}
\end{figure}

\autoref{fig:polluted_token_coverage} shows that pairwise overlap depends on the types of corpora being compared. Cleaner or more curated Chinese corpora tend to overlap substantially with one another: MAPCC and SkyPile cover more than half of each other's polluted tokens (similar for WanJuan, MAPCC, SkyPile, CCI3, and WuDao). This suggests that \textit{\textbf{cleaning and curation strategies still miss a shared set of residual polluted tokens}}, revealing a common limitation of existing filtering pipelines. By contrast, broader web corpora show more asymmetric containment. For example, 69.2\% of CulturaX polluted tokens appear in mC4, but only 13.8\% of mC4 polluted tokens appear in CulturaX. This suggests that \textit{\textbf{some broad web corpora subsume the polluted tokens of other corpora while also introducing a much larger additional tail}}.

\begin{figure}[t]
    \centering
    \includegraphics[width=0.95\linewidth]{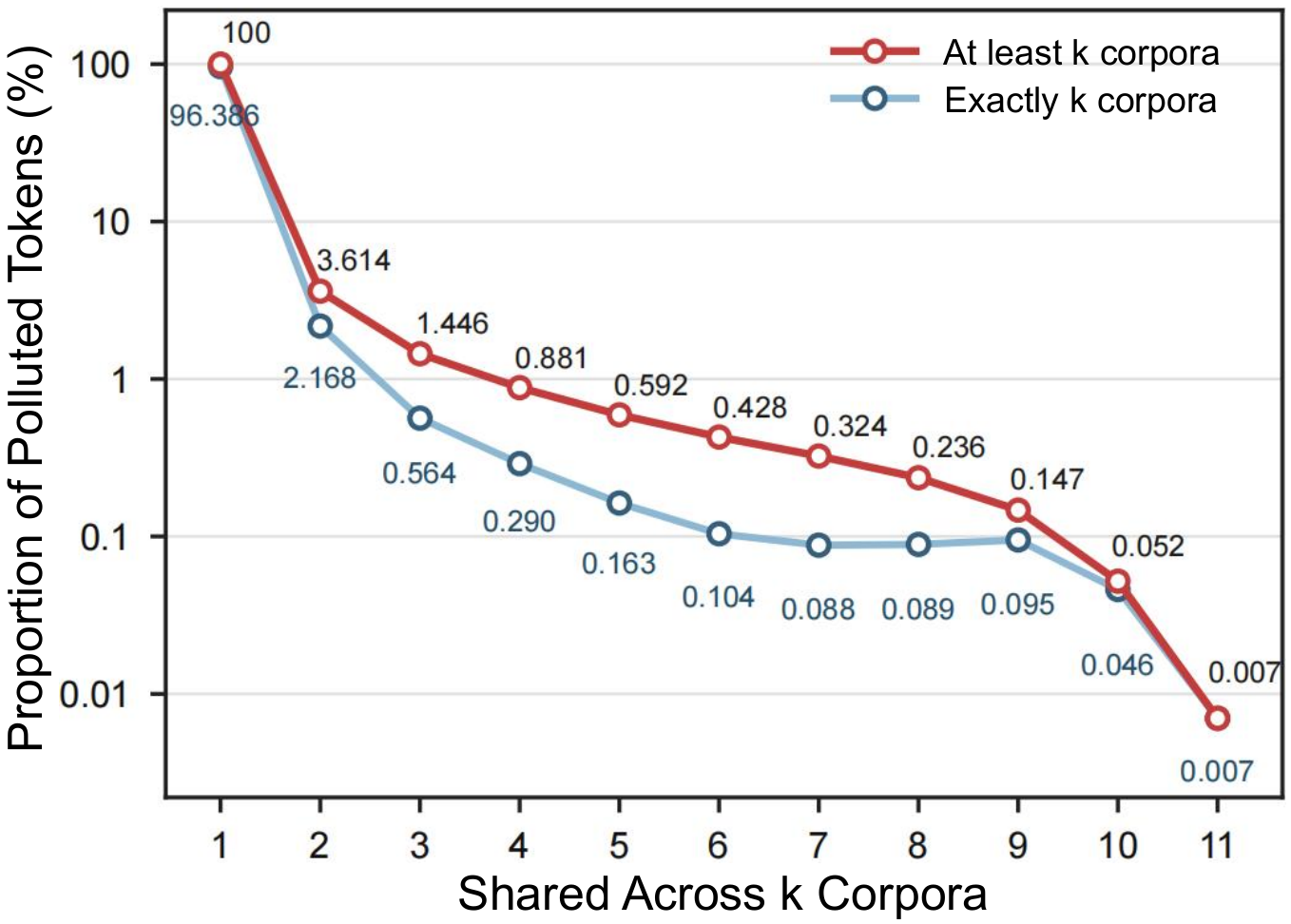}
    \caption{Proportion of polluted tokens shared across corpora. Blue line show the proportion appearing in exactly $k$ corpora, and the red line shows the cumulative proportion appearing in at least $k$ corpora.}
    \label{fig:polluted_token_dataset_count}
    \vspace{-1ex}
\end{figure}

To measure sharing beyond pairwise views, \autoref{fig:polluted_token_dataset_count} aggregates over all polluted tokens by counting how many corpora each token appears in. Across the 11 corpora, we observe 106{,}671 polluted tokens. Among them, 102{,}816 (96.386\%) appear in only one corpus. Only 1{,}542 (1.446\%) appear in at least three. This shows that \textit{\textbf{most polluted tokens are corpus-specific, with a small shared core recurring across multiple datasets}}.

\begin{figure}[t]
    \centering
    \includegraphics[width=\linewidth]{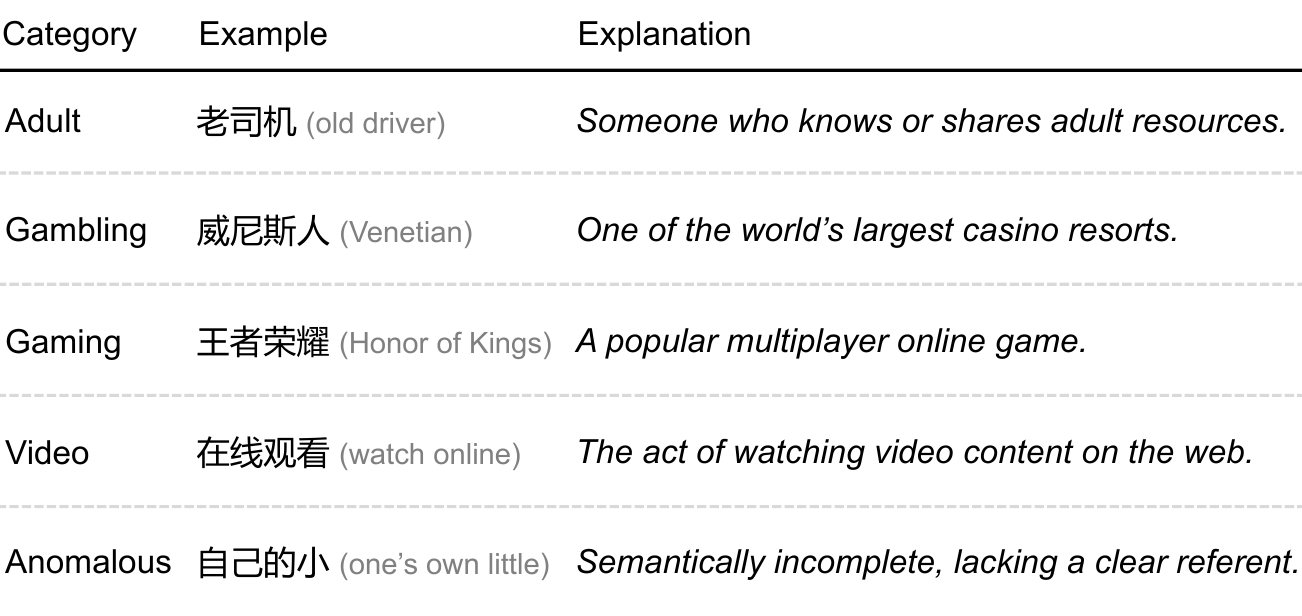}
    \caption{Representative examples from the shared core of polluted tokens. The shared core contains tokens that appear in at least 8 of the 11 corpora in \autoref{tab:zh_corpora_pollution}.}
    \label{fig:shared_tokens_corpora}
    \vspace{-1ex}
\end{figure}

\autoref{fig:shared_tokens_corpora} lists representative examples of the small shared core. 252 polluted tokens meet this criterion, yet they span all five polluted categories. These tokens are shared because they reflect long-lived lexical anchors. For example, ``老司机 (old driver)'' is a persistent euphemism in adult-resource pages, referring someone who knows or shares adult content resources; ``威尼斯人 (Venetian)'' is repeatedly reused as one of the world's largest casino resorts. In contrast, \autoref{app:dataset-specific-tokens} lists corpus-specific polluted tokens.

Together, these results show that \textit{\textbf{polluted tokens form a small shared core with much larger corpus-specific tails.}}

\section{Evolution of Chinese Web Content}
\label{sec:evolution}

The previous section shows that polluted tokens appear, to varying degrees, in mainstream open Chinese corpora. Since many of these corpora are directly or indirectly derived from Common Crawl \citep{commoncrawl,xue2021mt5,abadji2022oscar,nguy2024culturax,penedo2023refinedweb}, we next ask what the upstream source looks like: \textit{\textbf{How polluted is Chinese Common Crawl itself, and how does this pollution evolve over time?}}

Common Crawl is updated monthly, with each release containing web text at terabyte scale. For this temporal case study, we audit six Chinese Common Crawl snapshots from 2021 to 2026. For each snapshot, we run the same \textsc{Sampled-BPE} auditing pipeline with 1\% sampling rate, which yields lower than 3\% relative error according to \autoref{fig:sampling_accuracy}. Snapshot-selection and extraction details are provided in \autoref{app:chinese-corpora}.


\subsection{Pollution Is High and Shifting}

\autoref{tab:evolution_ratio} shows that Chinese Common Crawl contains strikingly high pollution, with Adult Content as the dominant source in several years. In 2026, total pollution even reaches 79.52\%, with Adult Content alone accounting for 68.72\%. This indicates that pollution in the upstream Chinese web source is not a marginal tail phenomenon, but becomes comparable to or even exceeds normal content.

\begin{table}[t]
\centering
\scriptsize
\resizebox{\linewidth}{!}{
\begin{tabular}{lrrrrrr}
\toprule
 & 2021 & 2022 & 2023 & 2024 & 2025 & 2026 \\
\midrule
\textbf{Pollution} & \textbf{35.35} & \textbf{61.87} & \textbf{62.90} & \textbf{34.83} & \textbf{38.01} & \textbf{79.52} \\
Adult & 24.73 & 54.53 & 55.86 & 28.01 & 30.53 & 68.72 \\
Gambling & 5.31 & 1.05 & 0.54 & 1.48 & 0.74 & 0.21 \\
Gaming & 0.40 & 0.16 & 0.23 & 0.33 & 0.22 & 0.07 \\
Video & 3.93 & 4.75 & 4.89 & 3.86 & 5.05 & 8.39 \\
Anomalous & 0.98 & 1.38 & 1.37 & 1.16 & 1.47 & 2.12 \\
\bottomrule
\end{tabular}
}
\caption{Temporal pollution ratios (\%) in Chinese Common Crawl snapshots. Chinese web content is highly polluted, and its pollution profile shifts over time}
\vspace{-2ex}
\label{tab:evolution_ratio}
\end{table}

Besides, different pollution categories evolve in different directions. Adult Content is both high and volatile, dropping from 55.86\% in 2023 to 28.01\% in 2024 before rising to 68.72\% in 2026. Online Gambling follows the opposite direction, decreasing from 5.31\% in 2021 to 0.21\% in 2026, a trend that is consistent with intensified enforcement against cross-border gambling activities targeting Chinese users\footnote{\url{https://english.scio.gov.cn/pressroom/2025-02/07/content_117699840.html}}. Online Video gradually increases from 3.93\% to 8.39\%, while Anomalous rises modestly and Online Gaming remains consistently small. These trends show that \textit{\textbf{Chinese web content is highly polluted, and its pollution profile shifts over time}}. Appendix word clouds in \autoref{app:wordclouds} further visualize the changing surface vocabulary across Common Crawl snapshots.

\subsection{Some Tokens Turn Over, Others Persist}

We further examine the evolution of tokens. For each category, we compute Jaccard overlap of tokens for adjacent years and for the five-year span, which captures how tokens turn over and persist.

\begin{figure}[t]
    \centering
    \includegraphics[width=\linewidth]{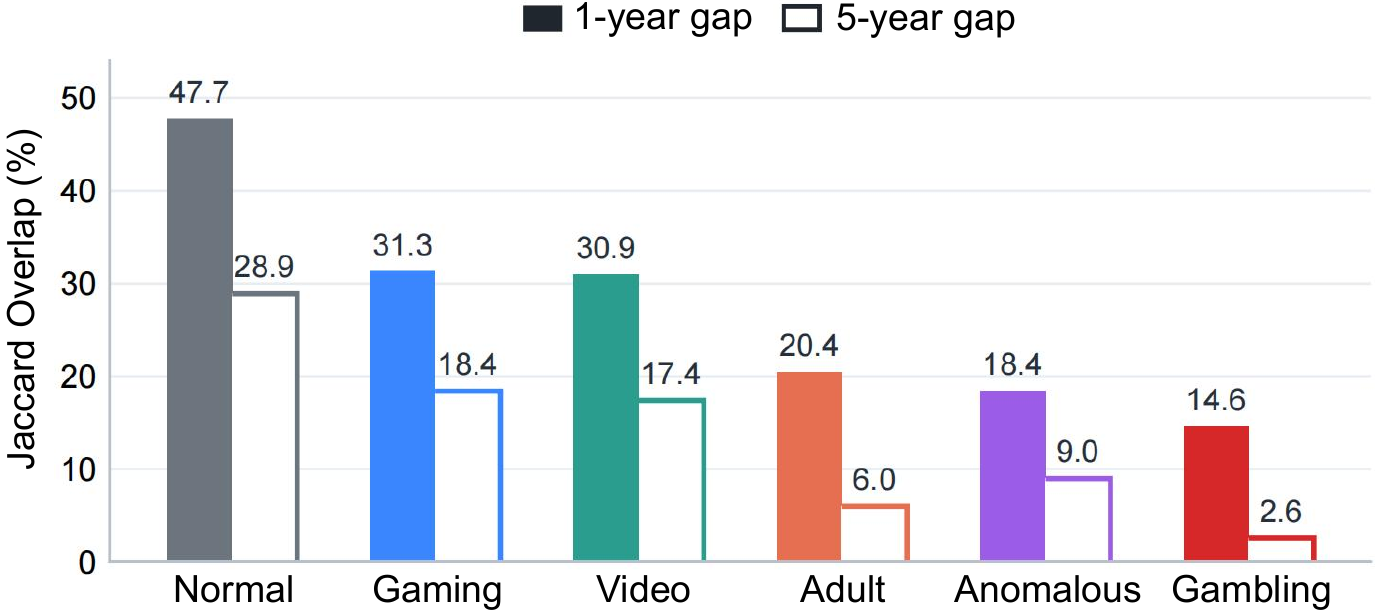}
    \caption{Jaccard overlap of tokens across years. Filled bars show average adjacent-year overlap, and hollow bars show the five-year (2021--2026) overlap.}
    \label{fig:evolution_main}
\end{figure}

\autoref{fig:evolution_main} shows that tokens in Normal Content, Online Gaming, and Online Video are replaced less frequently over time. Their adjacent-year overlap is relatively high, and their five-year overlap also remains non-trivial. This means these categories have less year-to-year token turnover and more long-lived tokens, suggesting that \textit{ordinary web content, online gaming, and online video are expressed through more stable forms on the web}.

By contrast, Adult Content, Online Gambling, and Anomalous show lower adjacent-year overlap and much weaker five-year persistence. The low adjacent-year overlap indicates frequent short-term token turnover, while the low five-year overlap indicates that few tokens persist over longer horizons. This may suggest that \textit{Adult Content and Online Gambling sites continually reappear on the Chinese web and change their surface forms even after being repeatedly banned or blocked, while Anomalous reflects the fast-changing vocabulary of subcultural or irregular web expressions} (see examples in \autoref{app:temporal-evolution}).

\begin{figure}[t]
    \centering
    \includegraphics[width=0.99\linewidth]{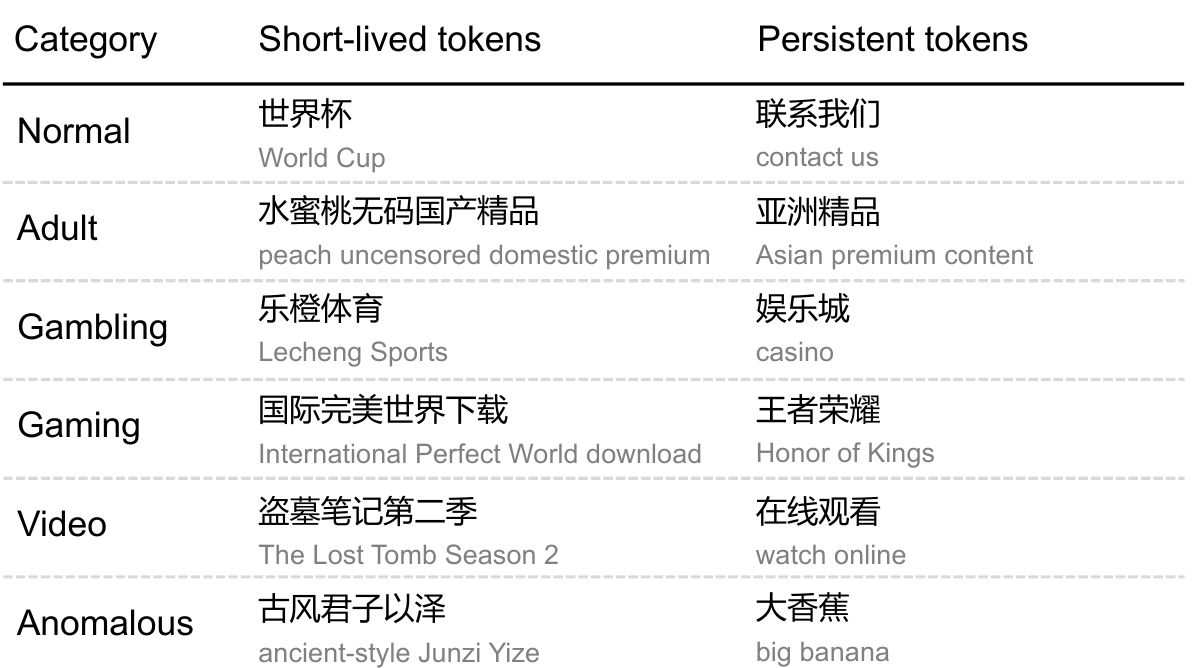}
    \caption{
    Representative tokens illustrating turnover and persistence in Chinese Common Crawl snapshots. 
    }
    \label{fig:evolution_token_examples}
    \vspace{-2ex}
\end{figure}

\autoref{fig:evolution_token_examples} provides representative tokens to interpret the contrast between stable and fast-changing categories. In the more stable categories such as Normal Content, Online Gaming, and Online Video, persistent tokens are reusable templates or stable entry points, such as ``联系我们 (contact us)'' and ``王者荣耀 (Honor of Kings)''. Their short-lived tokens tend to be concrete events, titles, or platform-specific phrases rather than broad changes in the category's surface form.

\begin{figure*}[t]
    \centering
    \includegraphics[width=\textwidth]{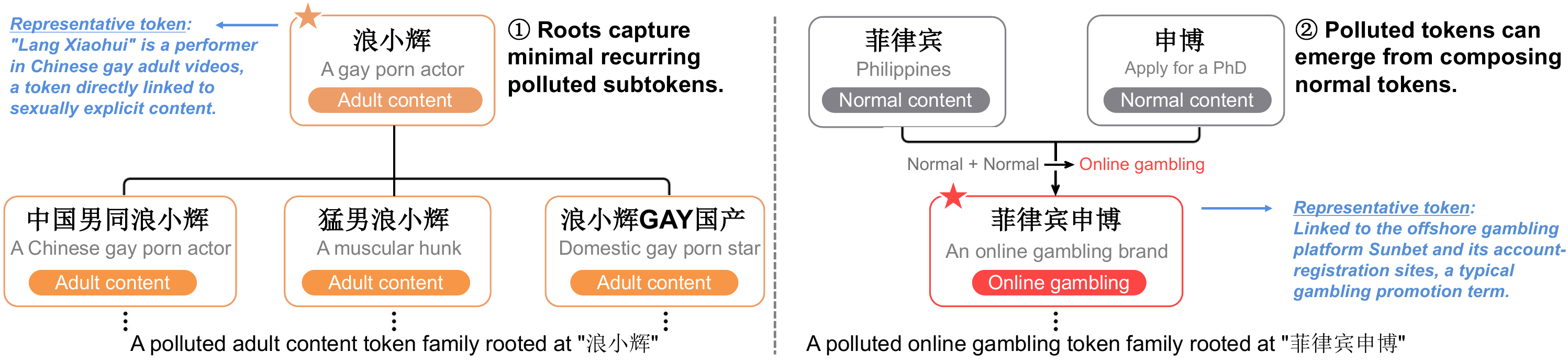}
    \caption{Examples of token trees in the hierarchical Chinese web token dataset. \textbf{Left}: roots identify minimal recurring subtokens of longer surface variants. \textbf{Right}: two normal subtokens can emerge pollution from combination.}
    \label{fig:token_trees}
    \vspace{-3ex}
\end{figure*}

In the faster-changing categories, short-lived tokens more often include adult-site phrases, gambling platforms, or subcultural expressions, such as ``水蜜桃无码国产精品 (peach uncensored domestic premium)'', ``乐橙体育 (Lecheng Sports)'', and ``古风君子以泽 (ancient-style Junzi Yize)''. Their persistent tokens mostly capture recurring templates, such as ``亚洲精品 (Asian premium content)'' and ``娱乐城 (casino)''. 

Thus, Common Crawl's Chinese pollution exhibits both category-level temporal drift and token-level lexical renewal. Static keyword lists or one-time filtering strategies can therefore quickly become stale, while \textbf{\textit{\textsc{Sampled-BPE} supports periodic auditing of upstream web corpora}}.

\section{Hierarchical Chinese Web Tokens}
\label{sec:hierarchical-tokens}

Building on our token-level audits of Chinese open corpora and Common Crawl, we release a hierarchical Chinese web token dataset with 630{,}684 tokens, so that these token-level findings can be reviewed, reused, and extended.

\subsection{Tokens Are Organized as Trees}

Chinese web tokens contain many substring relations. A short token may appear as a subtoken inside many longer tokens, and these longer tokens often form related surface variants. Instead of releasing a flat token list, we organize tokens as a forest (shown in \autoref{fig:token_trees}), consisting of 92{,}972 trees. \textbf{\textit{Each tree is rooted at the shortest subtoken in the family, and its descendants are longer tokens that contain the root}}.

Each token node contains Internet search evidence and a category label. Each tree further includes representative tokens and explanations for their categories. The dataset is therefore not only a collection of token strings, but also an auditable resource that records the contextual evidence and classification rationale.

\subsection{Hierarchy Traces Pollution}

We give two ways to trace pollution leveraging this hierarchical structure. 

First, \textit{when all tokens in a tree are from the same pollution family, the root identifies a minimal recurring subtoken}. This lets us summarize a set of surface variants with a shorter representative token, rather than treating every longer variant as an unrelated keyword. \autoref{fig:token_trees} (Left) shows an example of ``浪小辉 (A gay porn actor)'' which forms a large pollution family containing long variants such as ``浪小辉GAY国产''.

Second, \textit{trees with tokens from different categories reveal pollution that emerges from token composition}. A subtoken may be normal in isolation, but become polluted when combined with another normal token. \autoref{fig:token_trees} (Right) shows an example that ``菲律宾 (Philippines)'' and ``申博 (Apply for a PhD)'' can appear as normal tokens separately, while ``菲律宾申博 (An online gambling brand)'' is a gambling-related polluted token. Such cases are difficult to capture because the risk emerges from the composition rather than from either subtoken alone.

Overall, the hierarchy makes the dataset more informative than a flat list of polluted tokens. \textbf{\textit{It traces polluted token families back to minimal recurring subtoken and exposes hidden compositional cases.}} This provides evidence for future review, cleaning, and periodic auditing.

\section{Conclusion}

Motivated by Chinese web pollution surfacing in ChatGPT's vocabularies and Codex's outputs, this work presents \textsc{Sampled-BPE}, a lightweight token-level auditing pipeline that trains BPE tokenizers on sampled corpora to estimate Chinese corpora pollution profiles. Experiments show that \textsc{Sampled-BPE} substantially reduces runtime and memory while preserving usable estimates. 

We audit 11 open Chinese corpora and 6 Chinese Common Crawl snapshots. Results show that pollution is widespread but uneven across open Chinese corpora, and upstream Chinese web content is highly polluted and temporally shifting. A hierarchical Chinese web token dataset with 630k+ token records, each with web context, category, and explanation fields, is released for reviewing and tracing Chinese web pollution.

\clearpage
\section*{Limitations}

\paragraph{Pollution in other languages.}
This work focuses on polluted tokens in Chinese web-scale corpora and does not claim that the measured ratios transfer to other languages. \autoref{app:cross-lingual-pilot} reports a diagnostic Chinese--English boundary pilot, suggesting that English auditing likely requires phrase- or sentence-level modeling and language-specific evidence rather than direct transfer of the Chinese token-level setup.

\paragraph{Taxonomy scope.}
Our six categories follow prior work on polluted Chinese tokens \citep{zhang2025speculating} and are not intended as a universal data-safety taxonomy. Toxicity and political bias are usually context-dependent at the sentence level, while personally identifiable information is often compositional at the record level. Auditing these risks therefore requires complementary sentence- or document-level methods.

\paragraph{Readability to non-native Chinese readers.}
This paper necessarily includes many Chinese tokens, some of which are offensive, euphemistic, abbreviated, or context-dependent. We provide English glosses for representative examples when possible, but many tokens cannot be translated literally without losing their web-specific usage or pollution cues. Our goal is not to study Chinese linguistic expression itself, but to use Chinese web-scale corpora as a concrete case for studying upstream corpora pollution. We therefore treat translations as readability aids rather than complete semantic equivalents, and future work with broader multilingual expertise can improve these explanations.

\section*{Ethics Statement}
ACL Ethics Policy is respected in this work. This work studies pollution in open Chinese corpora and upstream Chinese Common Crawl snapshots. We use publicly available corpora and web evidence for research purposes, and we respect the terms, conditions, and copyright requirements of the corresponding data sources. Because the paper analyzes web pollution, it necessarily includes examples related to adult content, gambling, spam, anomalous expressions, and other potentially offensive or upsetting material. These examples are presented only for documenting and auditing corpus pollution, and should be used with caution in future research.

We adhere to the Association for Computational Linguistics (ACL) guidelines on responsible NLP research\footnote{\url{https://aclrollingreview.org/responsibleNLPresearch/}}, with particular attention to transparency, research-use framing, and responsible handling of harmful content.

\section*{Use of AI Assistants}
The authors used AI assistants for language polishing, LaTeX editing, and phrasing suggestions during paper preparation. All substantive claims, experimental results, analyses, citations, and final text were reviewed and verified by the authors.

\section*{Acknowledgements}
This work was supported by Alibaba Group through Alibaba Innovative Research Program.

\bibliography{custom}

\clearpage
\appendix

\section{Details of Open Chinese Corpora}
\label{app:chinese-corpora}

Following \autoref{sec:openChineseCorpora}, this section summarizes Common Crawl and the 11 representative open Chinese corpora audited in our experiments. For each corpus, we describe its source, public scale, audited Chinese portion when available, and relation to LLM training when the release is explicitly tied to a model. These details provide context for the corpus landscape in \autoref{fig:corpus} and the pollution ratios in \autoref{tab:zh_corpora_pollution}.

\autoref{tab:corpus-provenance} summarizes the audited releases and their upstream processing.

\begin{table*}[t]
\centering
\scriptsize
\setlength{\tabcolsep}{3pt}
\begin{tabular}{p{0.09\textwidth}p{0.20\textwidth}p{0.39\textwidth}p{0.24\textwidth}}
\toprule
\textbf{Corpus} & \textbf{Audited release/subset} & \textbf{Upstream processing status} & \textbf{Our audit protocol} \\
\midrule
HPLT & HPLT 3.0 \texttt{cmn\_Hans} & Extraction, LID, deduplication, quality scoring, metadata annotation, filtering & Audit the public Chinese portion \\
Common Crawl & February crawls 2021--2026; Chinese WET-derived records & Raw WET text; no single corpus-level cleaned release & Same Chinese extraction + 1\% sampling protocol across years; supports temporal comparison \\
OSCAR & OSCAR-2301-HPC Chinese portion & Derived from Common Crawl WET via Ungoliant & Audit the public Chinese portion \\
mC4 & mC4 Chinese files & C4-style multilingual cleaning/filtering & Audit the public Chinese files \\
CulturaX & CulturaX Chinese partition & mC4/OSCAR combination; LID, URL filtering, metric cleaning, document refinement, MinHash deduplication & Audit the public Chinese files \\
CWT & ChineseWebText full release; cleaner subset used in appendix comparison & EvalWeb deduplication, LID, rule filtering, BERT quality scoring; cleaner subset uses $>90$ quality score & Audit the public full/cleaner variants \\
ROOTS & ROOTS Chinese language group / audited Chinese slice & Curated multilingual corpus from OSCAR and documented sources & Audit the Chinese slice, with official denominator noted \\
WanJuan & Intern-WanJuan 1.0 text portion & Fine-grained cleaning, deduplication, safety-oriented filtering & Audit the public text release \\
MAPCC & MAP-CC \texttt{zh-cc} portion & Chinese-centric corpus including Common Crawl and curated sources & Audit the \texttt{zh-cc} portion \\
SkyPile & SkyPile-150B public plain-text portion & Filtering, deduplication, sensitive-data filtering, low-quality filtering & Audit the public plain-text portion \\
CCI3 & CCI3-HQ public files & Trusted Chinese Internet data; two-stage hybrid filtering & Audit the public high-quality subset \\
WuDao & WuDaoCorpus2.0 Base public subset & High-density page extraction; rule-based cleaning & Audit the public base subset \\
\bottomrule
\end{tabular}
\caption{Corpus provenance and upstream processing for audited corpora.}
\label{tab:corpus-provenance}
\end{table*}

\begin{figure}[t]
    \centering
    \includegraphics[width=\linewidth]{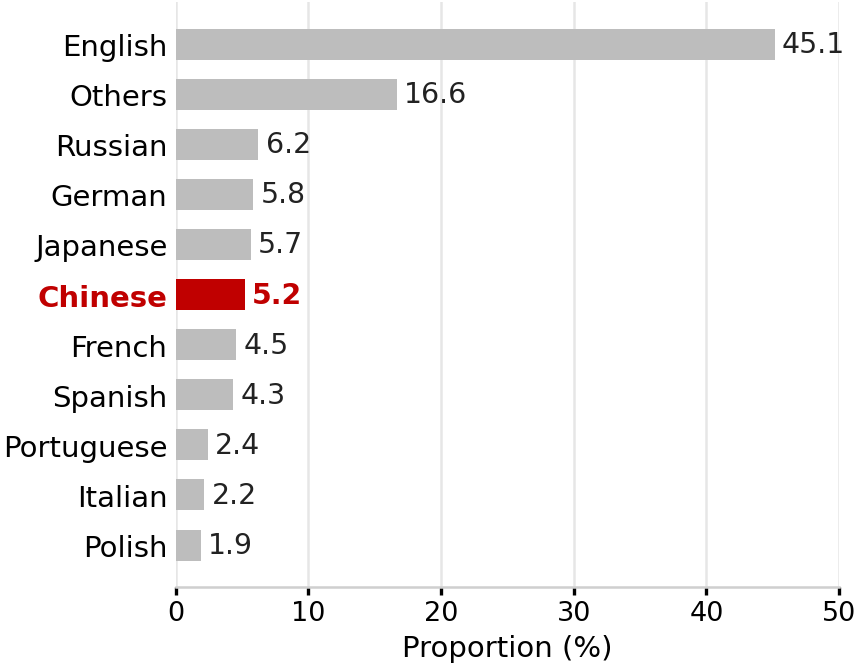}
    \caption{Language distribution in our Common Crawl WET accounting.}
    \label{appfig:cc-language-distribution}
\end{figure}

\noindent\textbf{CC} (Common Crawl Foundation, founded in 2007; data collected since 2008): Common Crawl is an open web-crawl archive maintained by a nonprofit organization. It has collected public web data since 2008 and, according to the official overview, now contains more than 10~PiB of archived data, with crawls published approximately once a month and each crawl typically containing more than two billion web pages. Common Crawl provides raw HTTP responses in WARC files, extracted metadata in WAT files, and extracted plaintext in WET files, making it the upstream source for many web-derived training corpora. In our WET-file accounting, each monthly crawl is approximately 6~TB compressed and 20~TB decompressed. Because Common Crawl does not publish official Chinese-only sizes, we estimate the language distribution by reading the WET header \texttt{WARC-Identified-Content-Language} and counting the first tag as the record's primary language; English accounts for 45.14\% (about 9~TB decompressed) and Chinese for 5.19\% (about 1~TB decompressed), as shown in \autoref{appfig:cc-language-distribution}. We audit Chinese WET-derived snapshots from 2021--2026 by selecting the February crawl in each year and sampling 1\% of the Chinese content, about 10~GB per crawl. Filtered or processed Common Crawl data has been used directly or indirectly in major models and corpora, including T5 through C4, GPT-3 through filtered Common Crawl, BLOOM through OSCAR/ROOTS, and Falcon through RefinedWeb~\citep{commoncrawl,raffel2020exploring,bigscience2022bloom,penedo2023refinedweb}.

\noindent\textbf{OSCAR} (OSCAR project, Inria/University of Mannheim/DFKI, 2023): OSCAR-2301-HPC is a document-oriented multilingual web corpus derived from Common Crawl WET files with the Ungoliant pipeline. The public OSCAR-2301-HPC table reports a Chinese portion of 138.48M documents, 44.38B space-separated words, and 1.4TB of content. OSCAR is intended as raw multilingual pretraining data; earlier OSCAR releases were also a major source in ROOTS, the corpus used to train BLOOM~\citep{abadji2022oscar,oscar2301hpc,laurencon2022roots,bigscience2022bloom}.

\noindent\textbf{mC4} (Google Research, 2021): Multilingual C4 is the multilingual variant of C4, built by cleaning Common Crawl text and extending the C4 filtering recipe to many languages. The public C4 release lists mC4 as a 9.7TB multilingual collection, and the Chinese files audited in this work total 71.42GB. mC4 was introduced as the pretraining corpus for mT5, a multilingual text-to-text Transformer covering 101 languages~\citep{xue2021mt5,allenai_c4}.

\noindent\textbf{HPLT} (European HPLT initiative, 2025): HPLT 3.0 is a very large multilingual web corpus built from Internet Archive and Common Crawl data from 2012--2024. HPLT 3.0 applies HTML extraction, language identification, deduplication, quality scoring, metadata annotation, and filtering; its Simplified Chinese Mandarin portion, \texttt{cmn\_Hans}, is reported as 5.40TB, 2.21B documents, 2.97T tokens, and 4.14T characters. The release is designed as a broad LLM and MT resource and is accompanied by HPLT-trained reference models and multilingual evaluations~\citep{oepen2025hplt3,hplt3_catalog}.

\noindent\textbf{CulturaX} (UONLP and collaborators, 2024): A cleaned multilingual dataset constructed by combining mC4 and OSCAR releases, followed by language identification, URL filtering, metric-based cleaning, document refinement, and MinHash deduplication. The full corpus contains 6.3T tokens in 167 languages; the Chinese partition contains 218.62M documents and 227.06B tokens, and the public Chinese files used for our audit total 641.02GB. CulturaX is released as reusable multilingual LLM training data rather than as a dataset paired with one flagship model~\citep{nguy2024culturax}.

\noindent\textbf{CWT} (CASIA-LM, 2023): ChineseWebText is a Chinese web corpus extracted with the EvalWeb toolchain from noisy Common Crawl data. EvalWeb combines deduplication, language identification, rule-based filtering, and a BERT-based quality evaluation model, and assigns a quality score to each text. The official release contains 1.42TB of scored Chinese web text and a cleaner 600GB subset with quality scores above 90\%, intended for Chinese LLM pretraining and threshold-based data selection~\citep{chen2023chinesewebtext}.

\noindent\textbf{ROOTS} (BigScience, 2022): The Responsible Open-science Open-collaboration Text Sources corpus is a 1.6TB multilingual collection spanning 59 languages, assembled from OSCAR and manually documented sources. The ROOTS Search Tool reports 259.01GB for the Chinese language group used as our official Chinese-size denominator. ROOTS was the training corpus for the 176B-parameter BLOOM multilingual language model~\citep{laurencon2022roots,rootssearch2023,bigscience2022bloom}.

\noindent\textbf{WanJuan} (Shanghai AI Laboratory/OpenDataLab, 2023): The text portion of Intern-WanJuan 1.0 is a cleaned pretraining corpus with more than 500M documents and more than 1TB of data. It integrates web pages, encyclopedias, books, patents, textbooks, and exam questions, converting heterogeneous HTML, text, PDF, and EPUB sources into a unified JSONL format after fine-grained cleaning, deduplication, and safety-oriented filtering. WanJuan was used in the training of InternLM~\citep{he2023wanjuan}.

\noindent\textbf{MAPCC} (M-A-P/CT-LLM team, 2024): MAP-CC is a Chinese-centric pretraining corpus released with the Chinese Tiny LLM project. MAP-CC contains about 800B Chinese tokens and is organized into components such as \texttt{zh-cc} from Chinese Common Crawl, encyclopedic data, papers, books, and other sources; the \texttt{zh-cc} portion audited here has a public size of 1.408TB. MAP-CC serves as the main Chinese pretraining data for CT-LLM, a 2B Chinese-centric language model trained together with English and code data~\citep{du2024chinesetinyllm,mapcc_dataset}.

\noindent\textbf{SkyPile} (Skywork, 2023): SkyPile-150B is a large-scale Chinese web dataset for pretraining language models. The public portion contains about 233M unique web pages, around 150B tokens, and 620GB of plain text. Its construction applies filtering, deduplication, sensitive-data filtering, and low-quality filtering with tools such as fastText and BERT. The release is associated with the Skywork bilingual foundation-model effort~\citep{wei2023skywork}.

\noindent\textbf{CCI3} (BAAI, 2024): CCI3-HQ is a high-quality subset of Chinese Corpora Internet 3.0 designed for LLM pretraining. The release is described as an approximately 500GB Chinese corpus, and the public files used for our audit total 518GB. It is produced from trusted Chinese Internet data using a two-stage hybrid filtering pipeline. The technical report evaluates the dataset by training a 0.5B model from scratch on 100B tokens and comparing against CCI3.0, SkyPile, and WanJuanV1~\citep{wang2024cci30hq}.

\noindent\textbf{WuDao} (Beijing Academy of Artificial Intelligence, 2021): WuDaoCorpus2.0 Base is the public base subset of WuDaoCorpora, a large-scale Chinese corpus collected for pretraining. The full WuDaoCorpora is described as a 3TB Chinese corpus, while the publicly released WuDaoCorpus2.0 Base contains 200GB of plain text. It is derived from large-scale Chinese web pages and other sources, with text extracted from high-density pages and cleaned through a rule-based pipeline. It is commonly used as a Chinese pretraining baseline and is associated with the WuDao large-model line~\citep{yuan2021wudaocorpora,wudaocorpusbase}.

\section{Robustness of Category Mapping}
\label{app:category-robustness}

We test whether category mapping depends on a particular search configuration or classifier backbone. All variants are evaluated on the same expert-annotated held-out split: \autoref{tab:search-robustness} compares search backends and regions, while \autoref{tab:backbone-robustness} compares classifier backbones.

\begin{table}[H]
\centering
\small
\begin{tabular}{lr}
\toprule
\textbf{Search setting} & \textbf{Held-out accuracy (\%)} \\
\midrule
Google / China region & 97.32 \\
Baidu API & 96.30 \\
Google / USA region & 96.56 \\
Google / Japan region & 96.81 \\
Google / Germany region & 96.68 \\
Google / Brazil region & 96.30 \\
\bottomrule
\end{tabular}
\caption{Robustness across search backends and regions.}
\label{tab:search-robustness}
\end{table}

\begin{table}[H]
\centering
\small
\begin{tabular}{lr}
\toprule
\textbf{LLM backbone} & \textbf{Held-out accuracy (\%)} \\
\midrule
GLM-4-32B & 97.32 \\
Qwen-2.5-72B & 95.64 \\
\bottomrule
\end{tabular}
\caption{Robustness across classifier backbones.}
\label{tab:backbone-robustness}
\end{table}

As \autoref{tab:search-robustness} shows, the search variants remain within 1.02 percentage points of the Google/China setting. \autoref{tab:backbone-robustness} further shows that Qwen-2.5-72B remains close to GLM-4-32B. Thus, mapping accuracy is not tied to one tested search configuration or backbone.

To measure how residual classification errors affect corpus-level estimates, we perturb token labels at the 2.68\% held-out error rate and recompute total pollution for each Common Crawl snapshot. The resulting estimates are reported in \autoref{tab:classification-error-propagation}.

\begin{table}[H]
\centering
\small
\setlength{\tabcolsep}{5pt}
\begin{tabular}{lrrr}
\toprule
\textbf{Year} & \textbf{Original} & \textbf{Perturbed} & \textbf{$\Delta$ (pp)} \\
\midrule
2021 & 35.35 & 36.89 & +1.54 \\
2022 & 61.87 & 62.56 & +0.69 \\
2023 & 62.90 & 63.56 & +0.66 \\
2024 & 34.83 & 36.38 & +1.55 \\
2025 & 38.01 & 39.46 & +1.45 \\
2026 & 79.52 & 79.64 & +0.12 \\
\bottomrule
\end{tabular}
\caption{Impact of 2.68\% label perturbation on total pollution ratios (\%). $\Delta$ is the absolute change from \autoref{tab:evolution_ratio} in percentage points.}
\label{tab:classification-error-propagation}
\end{table}

As \autoref{tab:classification-error-propagation} shows, the largest change is 1.55 percentage points, leaving the corpus-level scale and temporal conclusions unchanged.

\section{Chinese--English Comparison}
\label{app:cross-lingual-pilot}

To examine whether the same token-level discovery setup transfers beyond Chinese, we compare the full vocabularies of 150K-token BPE models trained on Chinese and English Common Crawl snapshots. As \autoref{tab:cross-lingual-boundary} shows, Chinese BPE directly surfaces complete polluted phrases and templates, whereas English BPE mainly surfaces isolated words or domain fragments.

\begin{table}[H]
\centering
\footnotesize
\setlength{\tabcolsep}{5pt}
\begin{tabularx}{\columnwidth}{@{}>{\raggedright\arraybackslash}p{0.22\columnwidth}>{\raggedright\arraybackslash}X@{}}
\toprule
\textbf{Chinese} & \textbf{Count: 76{,}404} \\
\textit{Examples} & ``成人av在线'' (``adult AV online''); ``无码免费视频'' (``uncensored free video''); ``{\CJKfamily{bsmi}線上娛樂城}'' (``online casino'') \\
\textit{BPE unit} & Complete polluted phrases or templates \\
\midrule
\textbf{English} & \textbf{Count: 413} \\
\textit{Examples} & \texttt{porn}; \texttt{casino}; \texttt{NakedCams} \\
\textit{BPE unit} & Isolated words or no-space domain fragments, rather than complete multi-word pollution units \\
\bottomrule
\end{tabularx}
\caption{Diagnostic comparison of the units surfaced by Chinese and English BPE vocabularies. Counts follow the language-specific checks used in this pilot.}
\label{tab:cross-lingual-boundary}
\end{table}

The smaller English count should not be interpreted as lower pollution prevalence. It instead marks a methodological boundary: English pollution is often expressed as multi-word phrases and therefore requires phrase- or sentence-level discovery rather than direct transfer of the Chinese token-level setup.

\section{Streaming vs. Random Sampling}
\label{app:streaming-vs-random}

Following \autoref{sec:auditing-pipeline}, this section compares the prefix-style streaming sampler used in \textsc{Sampled-BPE} with an index-based random sampler on the same CC reference setting. The random sampler first materializes a shuffled index set and then reads the selected rows, whereas the streaming sampler reads a contiguous prefix and can stop once the requested sample size has been reached. We use the 0.10\% sample size relative to the 30\% CC reference as a representative setting. In \autoref{apptab:streaming-vs-random-sampling}, arrows in the Streaming row show relative changes from Random.

\begin{table}[H]
\centering
\scriptsize
\resizebox{\linewidth}{!}{
\begin{tabular}{@{}lccc@{}}
\toprule
 & Time Cost & Token Coverage & Category WRE \\
\midrule
Random & 1731.7s & 70.19\% & 2.65\% \\
Streaming & 77.4s \diffdown{22.4\times} & 69.79\% \diffdown{0.57\%} & 2.57\% \diffdown{2.69\%} \\
\bottomrule
\end{tabular}
}
\caption{Streaming and random sampling comparison on CC at 0.10\% sampling rate.}
\label{apptab:streaming-vs-random-sampling}
\end{table}

\autoref{apptab:streaming-vs-random-sampling} shows that streaming sampling substantially reduces sampling time while preserving audit quality. Compared with random sampling, streaming sampling is 22.4$\times$ faster, while token coverage changes only from 70.19\% to 69.79\%. The category weighted relative error is also comparable, decreasing slightly from 2.65\% to 2.57\%. These results support our use of streaming sampling as a low-cost system choice for web-scale auditing.

\section{Cleaning Reduces but Does Not Eliminate Pollution}
\label{app:chinese-cleaning}

Following \autoref{sec:open-corpus-audits}, this section provides focused comparisons supporting the claim that cleaning and curation reduce, but do not eliminate, Chinese web pollution. We compare ChineseWebText full and cleaner variants, and two ROOTS Chinese variants with different source and curation composition. Values follow the same normalization as \autoref{tab:zh_corpora_pollution}.

\begin{table}[H]
\centering
\scriptsize
\setlength{\tabcolsep}{5pt}
\begin{tabular}{lcc@{\hspace{1.2em}}cc}
\toprule
 & \multicolumn{2}{c}{ChineseWebText} & \multicolumn{2}{c}{ROOTS} \\
\cmidrule(lr){2-3}\cmidrule(lr){4-5}
 & Full & Cleaner & Uncorpus & Public \\
\midrule
\textbf{Pollution} & \textbf{2.35} & \textbf{0.50} & \textbf{2.02} & \textbf{1.37} \\
\addlinespace[1pt]
Adult & 0.05 & 0.05 & 0.23 & 0.19 \\
Gambling & 0.22 & 0.02 & 0.00 & 0.01 \\
Gaming & 0.29 & 0.05 & 0.00 & 0.01 \\
Video & 0.06 & 0.02 & 0.00 & 0.00 \\
Anomalous & 1.73 & 0.37 & 1.79 & 1.16 \\
\bottomrule
\end{tabular}
\caption{Pollution ratios (\%) for cleaned and less-cleaned Chinese corpus variants.}
\label{apptab:zh-cleaning}
\end{table}

\autoref{apptab:zh-cleaning} shows that cleaning sharply lowers pollution in ChineseWebText: total pollution drops from 2.35\% in CWT-Full to 0.50\% in CWT-Cleaner. The reduction is especially clear for Anomalous content, which decreases from 1.73\% to 0.37\%, and Online Gambling, which decreases from 0.22\% to 0.02\%. ROOTS shows a similar but weaker pattern: ROOTS-Public has lower total pollution than ROOTS-Uncorpus (1.37\% vs. 2.02\%), but still retains measurable residual pollution, mainly from Anomalous and Adult Content. These comparisons support the main-text claim that cleaning helps, but does not remove polluted tokens entirely.

\section{Sampling Rates of Chinese Corpora}
\label{app:sampling-rates}

Following \autoref{sec:open-corpus-audits}, this section reports the actual sampling rates used for auditing the 11 open Chinese corpora and the corresponding full-token weighted relative errors estimated from \autoref{fig:sampling_accuracy}. Sampling rates use official public corpus-size denominators where available. Weighted errors are obtained by log-linear interpolation over the full-token category weighted-error curve.

\begin{table*}[h!]
\centering
\scriptsize
\resizebox{\textwidth}{!}{
\begin{tabular}{@{}lccccccccccc@{}}
\toprule
Corpus & OSCAR & mC4 & HPLT & CulturaX & CWT & ROOTS & WanJuan & MAPCC & SkyPile & CCI3 & WuDao \\
\midrule
Sampling rate & 1.53\% & 14.00\% & 0.19\% & 0.94\% & 0.76\% & 0.46\% & 0.53\% & 0.36\% & 1.94\% & 2.03\% & 1.34\% \\
Weighted relative error & 2.34\% & 0.24\% & 3.63\% & 2.75\% & 2.93\% & 2.90\% & 3.02\% & 2.68\% & 2.06\% & 2.00\% & 2.46\% \\
\bottomrule
\end{tabular}
}
\caption{Sampling rates and estimated full-token weighted relative errors for auditing the 11 open Chinese corpora.}
\label{apptab:sampling-rates-weighted-error}
\end{table*}

For ROOTS, the denominator is the complete Chinese ROOTS size, not only the audited \texttt{roots\_zh\_uncorpus} slice. For WanJuan, the official denominator is reported only as larger than 1TB, so the sampling rate is an upper bound and the error is interpolated at 0.53\%.

\autoref{apptab:sampling-rates-weighted-error} shows that we use small sampling rates for most corpora, ranging from 0.19\% for HPLT to 2.03\% for CCI3, with mC4 as an exception because the audited Chinese slice is much smaller. Despite these low sampling rates, the estimated full-token weighted relative errors remain small: all are below 4\%, and most fall between 2\% and 3\%. This supports the result that \textsc{Sampled-BPE} can provide usable corpus pollution estimates without full-corpus auditing.

\section{Additional Token Evolution Analyses}
\label{app:temporal-evolution}

Following \autoref{sec:evolution}, this section supplements the main all-token temporal analysis with high-frequency top-100 token views and a top-$k$ sensitivity check. We focus on token turnover and persistence in Chinese Common Crawl snapshots, using two complementary signals: Jaccard overlap across time gaps and the average number of yearly entrants in the top-100 list.

\begin{figure}[t]
    \centering
    \includegraphics[width=\linewidth]{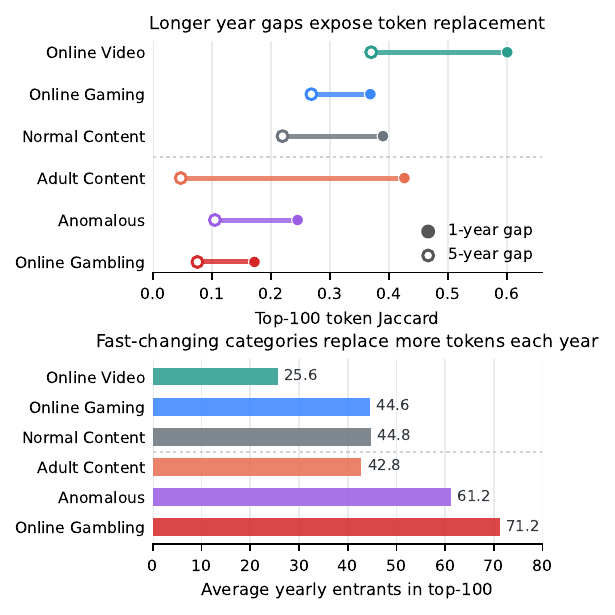}
    \caption{Additional top-100 token evolution signals across Chinese Common Crawl snapshots. \textbf{Top}: adjacent-year and five-year Jaccard overlap. \textbf{Bottom}: average yearly entrants in the top-100 lists.}
    \label{appfig:evolution-persistence-turnover}
\end{figure}

\autoref{appfig:evolution-persistence-turnover} shows a clear contrast between stable and fast-changing categories. Online Video has the highest adjacent-year overlap and the fewest yearly entrants, indicating that its high-frequency token head is comparatively stable. Online Gambling shows the opposite pattern: it has the lowest overlap and the largest number of yearly entrants, indicating rapid replacement of high-frequency tokens. Anomalous also shows substantial turnover, with many new top-100 tokens entering each year.

Adult Content is more nuanced. Its adjacent-year top-token Jaccard remains moderate, but its five-year overlap is much lower, suggesting that this category combines short-term template stability with longer-term lexical replacement. This explains why Adult Content still shows substantial temporal drift even when recurring fragments such as ``国产精品 (domestic premium content)'', ``精品视频 (premium videos)'', and ``精品国产 (domestic premium series)'' remain visible across snapshots.

\paragraph{Sensitivity to the top-$k$ threshold.}
We further repeat the lexical-evolution analysis with top-$10$, top-$100$, and full category vocabularies to test whether the conclusions depend on the truncation threshold. The main pattern is that smaller $k$ emphasizes stability in the lexical head, whereas the full vocabulary exposes volatility in the long tail. This effect is especially strong for Online Video, whose adjacent-year Jaccard drops from 0.76 at top-$10$ to 0.60 at top-$100$ and then to 0.31 over the full vocabulary. Adult Content and Anomalous show similar declines, from 0.53 to 0.43 to 0.20 and from 0.49 to 0.25 to 0.18, respectively. In contrast, Normal Content remains relatively stable across thresholds, with adjacent-year Jaccard of 0.39 at top-$10$, 0.39 at top-$100$, and 0.48 over the full vocabulary. Online Gambling is unstable at every scale, with adjacent-year Jaccard staying low from 0.24 to 0.17 and 0.15. Overall, these ablations motivate the main-text all-token view when the goal is to expose long-tail lexical renewal, while the top-$100$ analysis keeps the high-frequency patterns interpretable.

\paragraph{Examples of short-lived and long-lived tokens.}
\begin{packeditemize}
    \item \textbf{Normal Content.} Short-lived examples include ``影片名称 (film title)'', ``世界杯 (World Cup)'', and ``我欲封天 (I Shall Seal the Heavens)'', which spike in a single year; long-lived examples include ``中文字幕 (Chinese subtitles)'', ``联系我们 (contact us)'', ``有限公司 (limited company)'', and ``关于我们 (about us)'', which persist across most or all snapshots. \textbf{\textit{This category is comparatively stable overall.}}
    \item \textbf{Adult Content.} Short-lived examples include ``日本一级特黄大片 (Japanese explicit-content phrase)'', ``亚洲成人 (Asian adult)'', ``亚洲天堂 (Asian paradise)'', and ``一区二区在线观看 (Zone 1/2 watch online)''; long-lived examples include ``国产精品 (domestic premium content)'', ``精品视频 (premium videos)'', ``精品国产 (domestic premium series)'', and ``久久久久久 (repeated jiu-jiu template)'', reflecting a stable template-like core mixed with rapid surface turnover. \textbf{\textit{The key pattern is stable templates plus changing surface realizations.}}
    \item \textbf{Online Gambling.} Short-lived examples include ``太阳城 (Sun City)'', ``乐橙体育 (Lecheng Sports)'', ``申博太阳城 (Shenbo Sun City)'', and ``云顶集团手机版 (Genting Group mobile version)''; long-lived examples include ``开奖结果 (lottery results)'', ``双色球 (Double Color Ball)'', ``娱乐城 (casino)'', ``免费一级 (free first-level content)'', and ``澳门正版资料 (Macau official materials)'', showing recurring gambling-related templates but rapid platform and brand replacement. \textbf{\textit{This is the clearest case of rapid lexical turnover.}}
    \item \textbf{Online Video.} Short-lived examples include ``盗墓笔记第二季 (The Lost Tomb Season 2)'', ``完美世界txt下载 (Perfect World TXT download)'', and ``网址在线观看 (watch online at URL)''; long-lived examples include ``在线观看 (watch online)'', ``在线播放 (stream online)'', ``在线视频 (online video)'', ``免费视频 (free video)'', and ``免费观看 (watch for free)'', consistent with the high stability of generic video-access phrases. \textbf{\textit{This is the most stable category.}}
    \item \textbf{Anomalous.} Short-lived examples include ``小说零 (novel zero)'', ``一區二區 (Zone 1/Zone 2)'', ``五月婷 (May Ting)'', and ``最高占成 (highest commission share)''; long-lived examples include ``一区二区三区 (Zones 1/2/3)'', ``一区二区 (Zones 1/2)'', ``一二三区 (Zones 1/2/3)'', ``大香蕉 (big banana)'', and ``卡三卡 (card-three-card)'', illustrating both recurring peculiar phrases and rapidly changing surface variants. \textbf{\textit{This category combines contextually irrelevant recurring forms with strong year-to-year drift.}}
    \item \textbf{Online Gaming.} Short-lived examples include ``完美世界前传下载 (Perfect World prequel download)'', ``国际完美世界下载 (International Perfect World download)'', and ``小说改编的网页游戏 (web games adapted from novels)''; long-lived examples include ``破解版 (cracked version)'', ``小游戏 (mini games)'', ``游戏平台 (game platform)'', ``王者荣耀 (Honor of Kings)'', and ``游戏下载 (game download)'', indicating a more persistent gaming vocabulary than Gambling or Anomalous. \textbf{\textit{Its stability is closer to mainstream content than to Gambling or Anomalous.}}
\end{packeditemize}

\section{Corpus-Specific Polluted Tokens}
\label{app:dataset-specific-tokens}

Following \autoref{sec:token_level_pollution}, this section complements the shared-core analysis by showing representative polluted tokens that are unique to one audited corpus slice. These examples are grouped by predicted category.

\begin{figure*}[p]
  \centering
  \includegraphics[height=0.9\textheight]{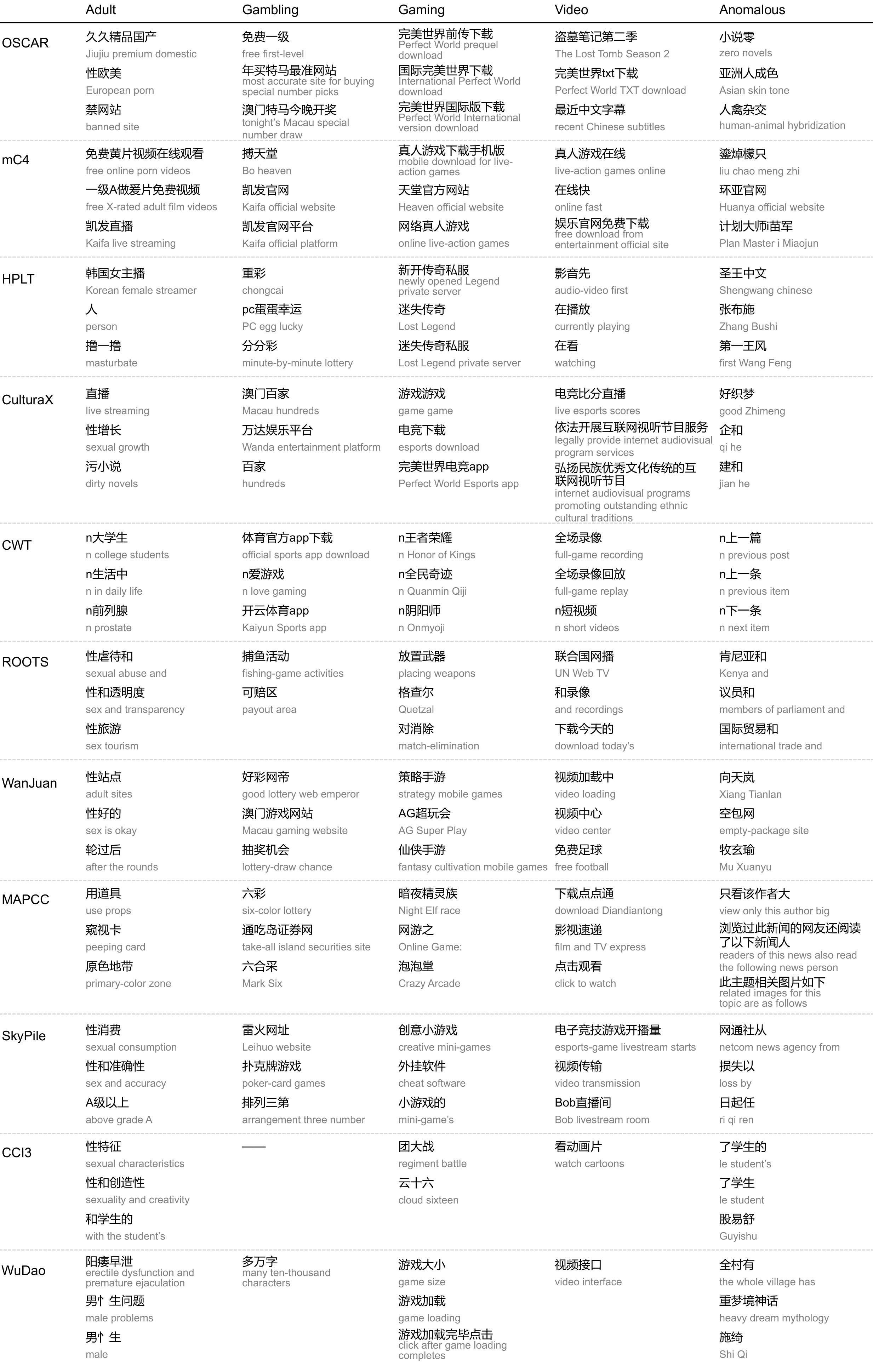}
  \caption{Representative dataset-specific polluted tokens, with English glosses.
           Each cell lists high-ratio tokens that appear in only one of the 11
           representative corpus slices, grouped by predicted category.}
  \label{fig:dataset-tokens}
\end{figure*}

\autoref{fig:dataset-tokens} shows that corpus-specific polluted tokens are not limited to a single pollution type. Different corpus slices contain distinct Adult Content, Online Gambling, Online Gaming, Online Video, and Anomalous tokens, reflecting source-specific and pipeline-specific lexical artifacts. This provides qualitative support that most polluted tokens form corpus-specific tails rather than a shared vocabulary across all corpora.


\section{Token Clouds}
\label{app:wordclouds}

Following \autoref{sec:open-corpus-audits} and \autoref{sec:evolution}, this section provides qualitative token-cloud views of high-frequency Chinese tokens in the audited open corpora and Chinese Common Crawl snapshots. For each corpus, we read long Chinese tokens (more than 2 Chinese characters), use token-count statistics to rank tokens, and visualize the top-2{,}000 tokens. Font size is determined by count ranking. These token clouds visualize surface forms rather than category ratios, providing qualitative context for \autoref{tab:zh_corpora_pollution} and \autoref{tab:evolution_ratio}.

\begin{figure*}[p]
    \centering
    \includegraphics[width=\textwidth]{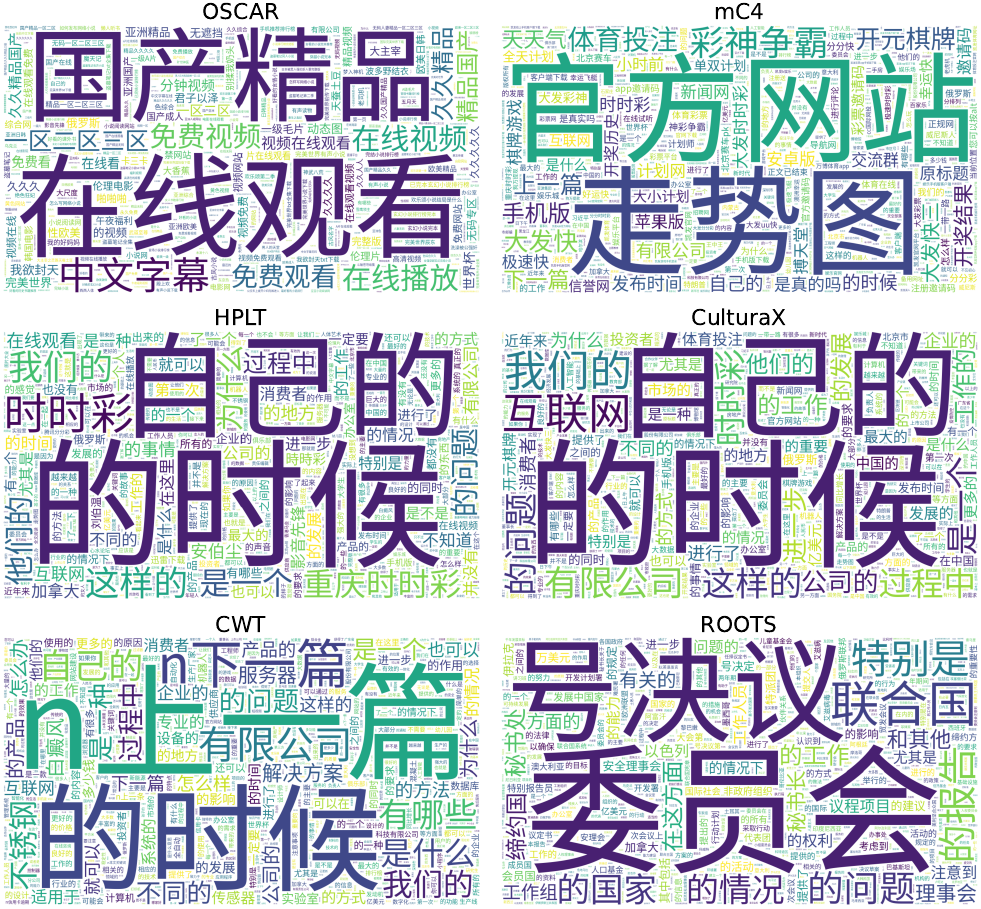}
    \caption{Token clouds for six representative broad-web corpus slices.}
    \label{appfig:wordcloud-broad-web-corpora}
\end{figure*}

\autoref{appfig:wordcloud-broad-web-corpora} shows that the six representative broad-web corpus slices differ sharply in their high-frequency surface forms. OSCAR is visually dominated by video and adult-access templates, whereas mC4 contains many lottery- or betting-like phrases. HPLT, CulturaX, CWT, and ROOTS show more generic web, institutional, or template-like text. This qualitative contrast aligns with the main result that pollution is widespread but uneven across corpus families, and that different corpora exhibit different dominant pollution profiles.

\begin{figure*}[p]
    \centering
    \includegraphics[width=\textwidth]{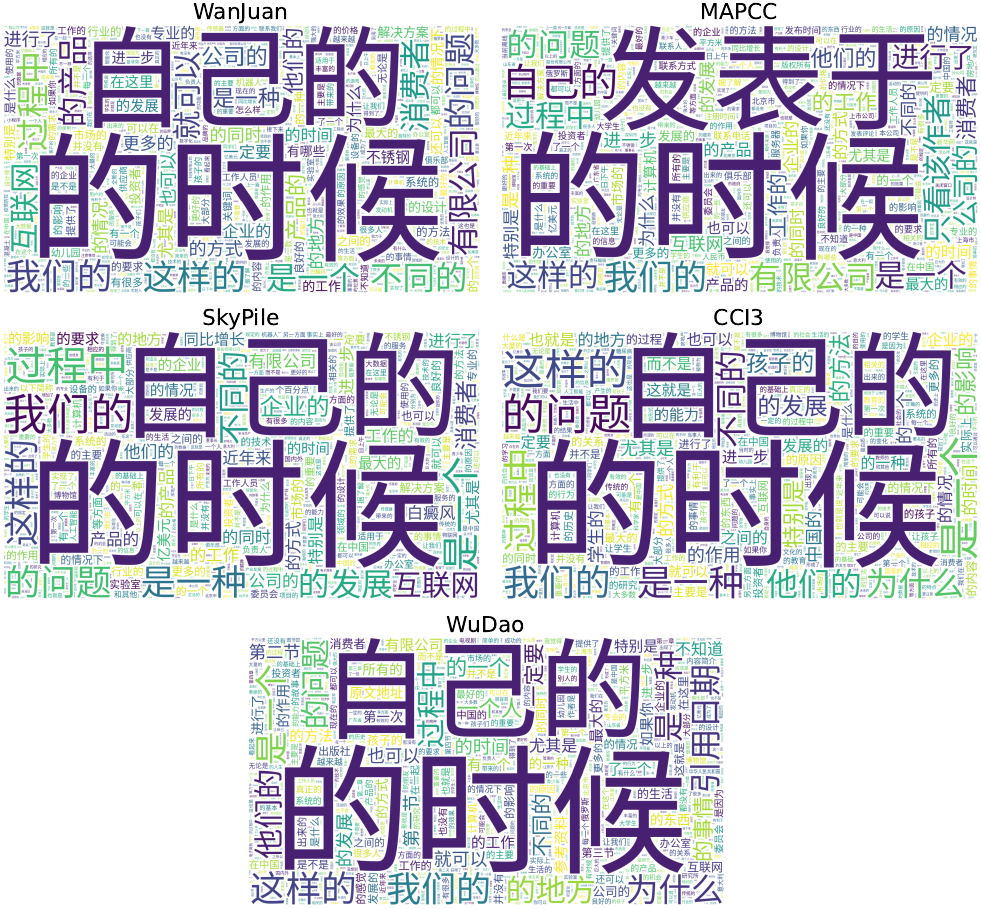}
    \caption{Token clouds for curated Chinese corpus slices.}
    \label{appfig:wordcloud-curated-chinese-corpora}
\end{figure*}

\autoref{appfig:wordcloud-curated-chinese-corpora} shows a different pattern for curated Chinese corpora. Their top tokens are mostly generic content or web-template phrases, consistent with their much lower pollution ratios in \autoref{tab:zh_corpora_pollution}. However, because token clouds only visualize the highest-count tokens, they should not be read as evidence that these corpora are pollution-free. Rather, they show that in cleaner corpora, residual polluted tokens are less visually dominant and often remain in a lower-frequency tail.

\begin{figure*}[p]
    \centering
    \includegraphics[width=\textwidth]{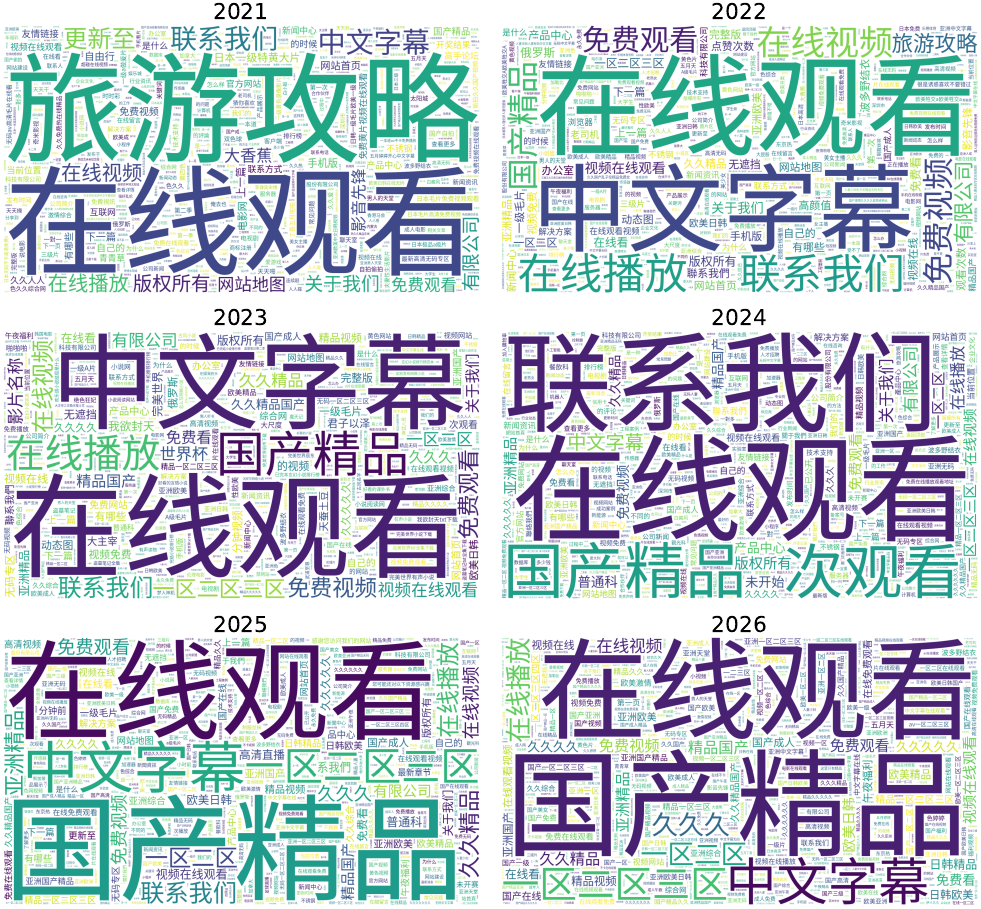}
    \caption{Token clouds for Chinese Common Crawl snapshots from 2021 to 2026.}
    \label{appfig:wordcloud-common-crawl-snapshots}
\end{figure*}

\autoref{appfig:wordcloud-common-crawl-snapshots} provides a temporal view of Common Crawl. The dominant high-frequency tokens change across yearly snapshots, with visible shifts among video-access phrases, site templates, and adult-content templates. This supports the temporal analysis in \autoref{sec:evolution}: Chinese Common Crawl pollution is not only high in aggregate, but also changes in its surface vocabulary over time.

\clearpage

\section{Additional Token Tree and Search Evidence Examples}
\label{app:token-tree-evidence}

Following \autoref{sec:hierarchical-tokens}, this section provides additional token-tree examples and their corresponding search evidence. These examples expand the main-text discussion in two directions. First, they show \textbf{\textit{how a tree can trace an entire polluted family back to a compact root}}, making related surface variants reviewable as one group rather than as disconnected strings. Second, they show \textbf{\textit{how pollution can arise compositionally}}: a longer token can become polluted even when the shorter subtokens from which it is formed are normal in isolation. The paired search-evidence figures further show why category labels are assigned with contextual web evidence rather than from token strings alone.

\begin{figure*}[h!]
  \centering
  \includegraphics[width=\textwidth]{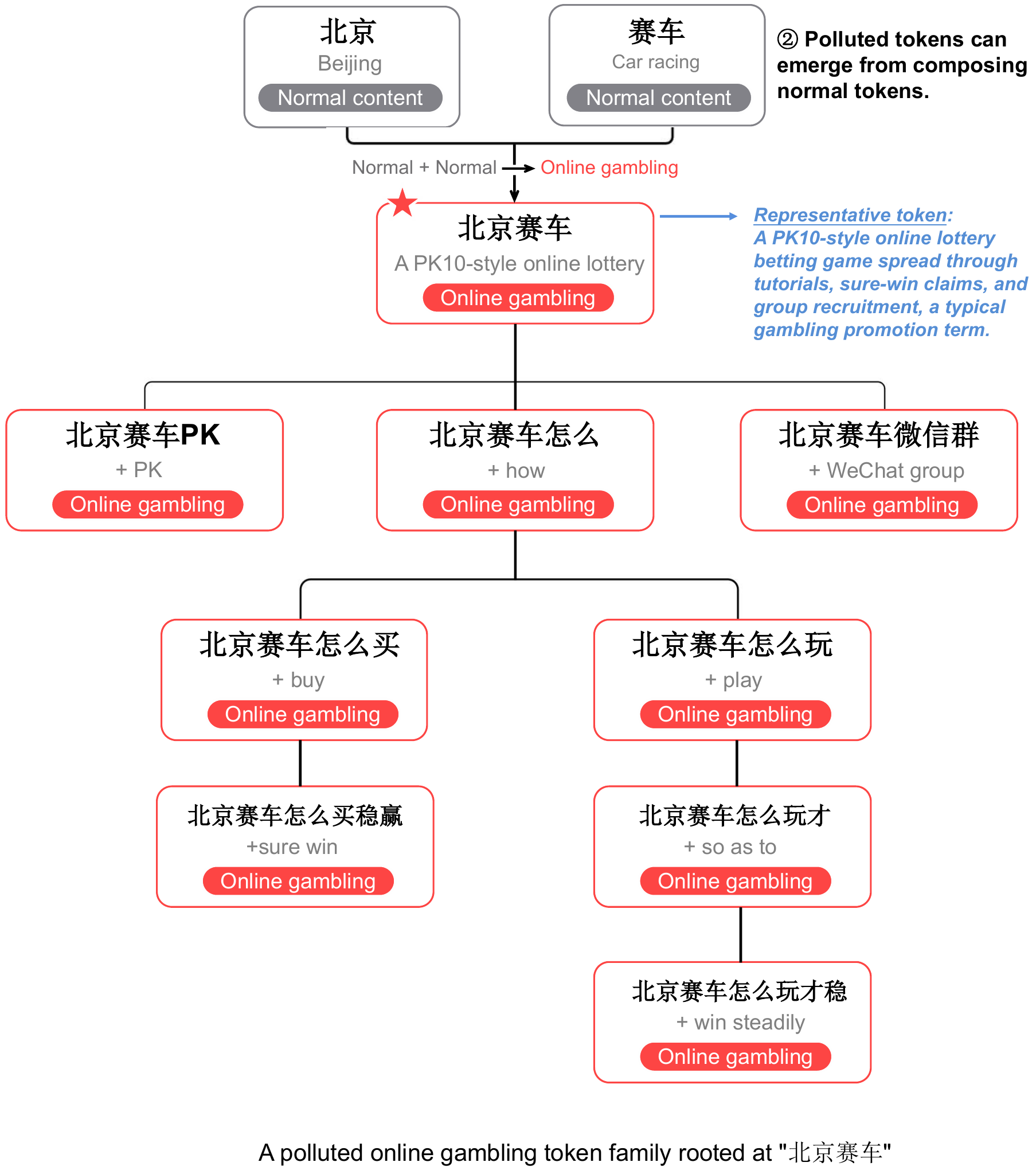}
  \caption{Token tree rooted at ``北京赛车''. The tree shows how two individually normal subtokens, ``北京'' and ``赛车'', compose into an Online Gambling token family.}
  \label{fig:token-tree-saiche}
\end{figure*}

\autoref{fig:token-tree-saiche} illustrates the compositional case with ``北京赛车''. The shorter subtokens ``北京 (Beijing)'' and ``赛车 (Car racing)'' are normal when interpreted separately, but their combination refers to a PK10-style lottery betting term. The descendants then capture common gambling-search variants such as platform, strategy, purchase, and WeChat-group phrases. This example shows that a flat keyword list would either miss the composed polluted expression or over-block its normal components; the tree instead localizes the polluted meaning to the composed token family.

\begin{figure*}[p]
  \centering
  \includegraphics[width=\textwidth]{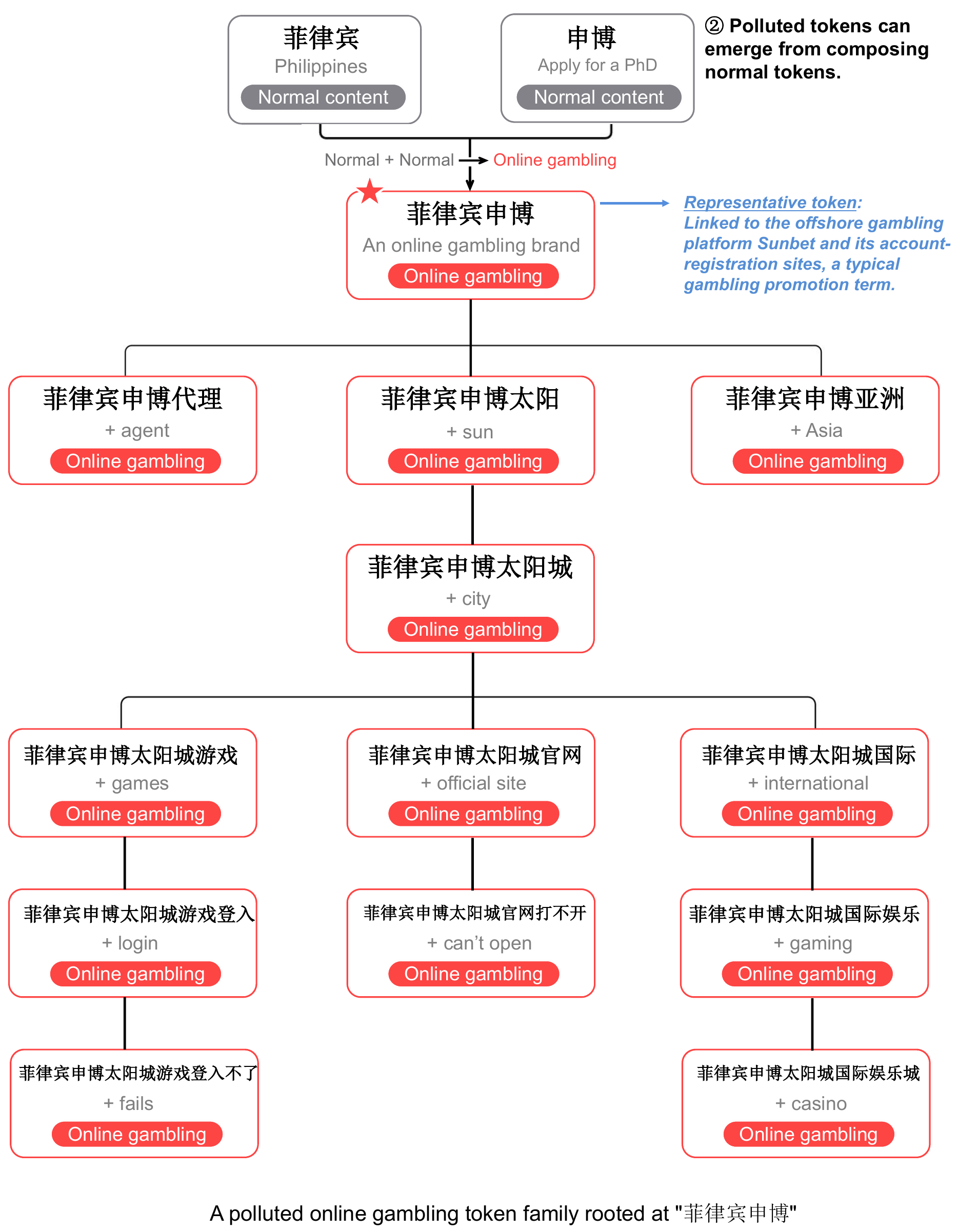}
  \caption{Token tree rooted at ``菲律宾申博''. The tree shows how normal subtokens can compose into a family of Online Gambling tokens.}
  \label{fig:token-tree-shenbo}
\end{figure*}

\autoref{fig:token-tree-shenbo} shows a second compositional case. Here ``菲律宾 (Philippines)'' and ``申博 (Apply for a PhD)'' can be normal tokens separately, but their combination is associated with an online gambling brand. Descendants such as agent, official-site, login, and international-casino variants reveal how one composed token expands into a broader promotional family. This case is important because it shows that pollution can emerge from ordinary-looking named entities and abbreviations, rather than from an obviously harmful root.

\begin{figure*}[p]
  \centering
  \includegraphics[width=\textwidth]{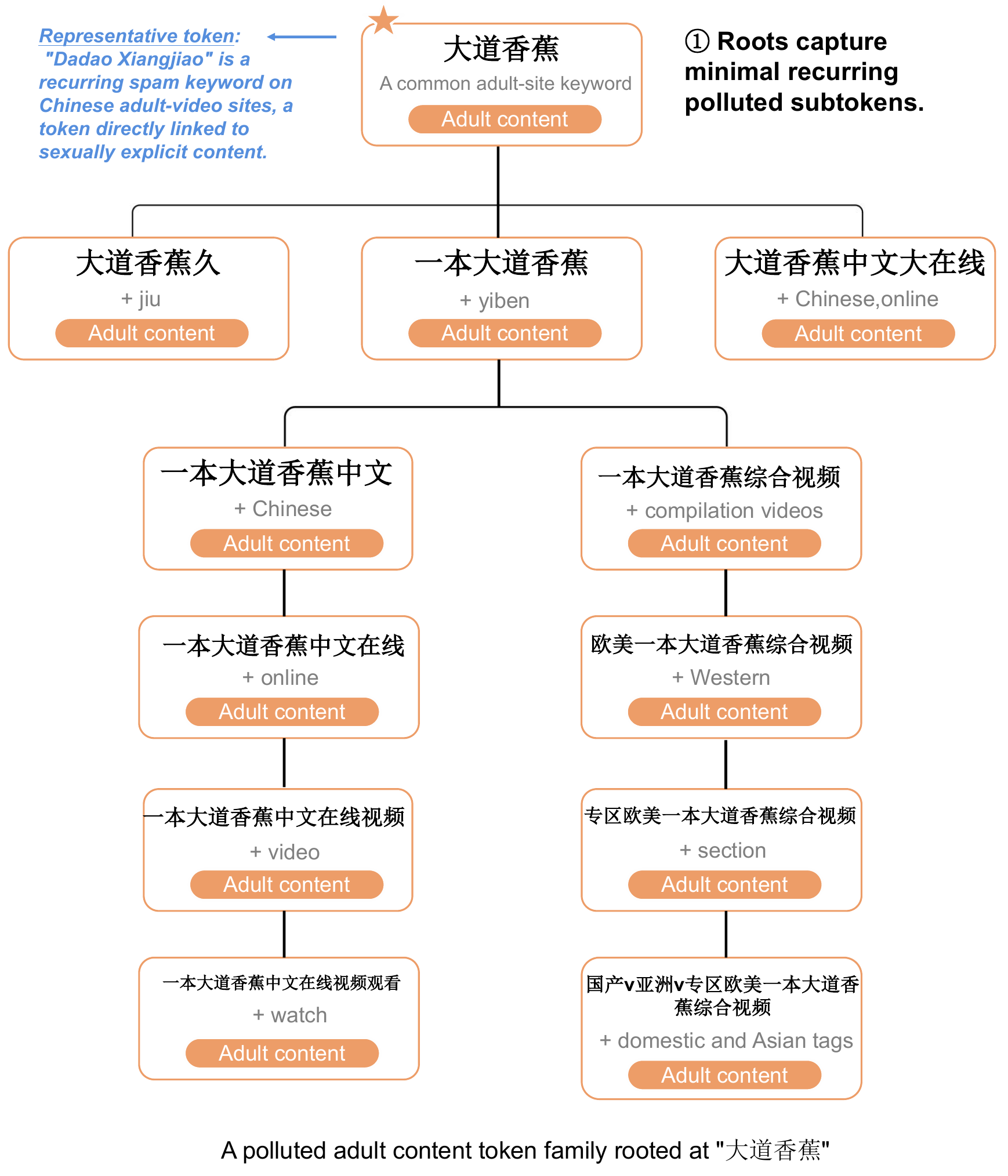}
  \caption{Token tree rooted at ``大道香蕉''. The root captures a recurring Adult Content subtoken, and descendants show longer adult-site variants.}
  \label{fig:token-tree-xiangjiao}
\end{figure*}

\begin{figure*}[p]
  \centering
  \includegraphics[width=\textwidth]{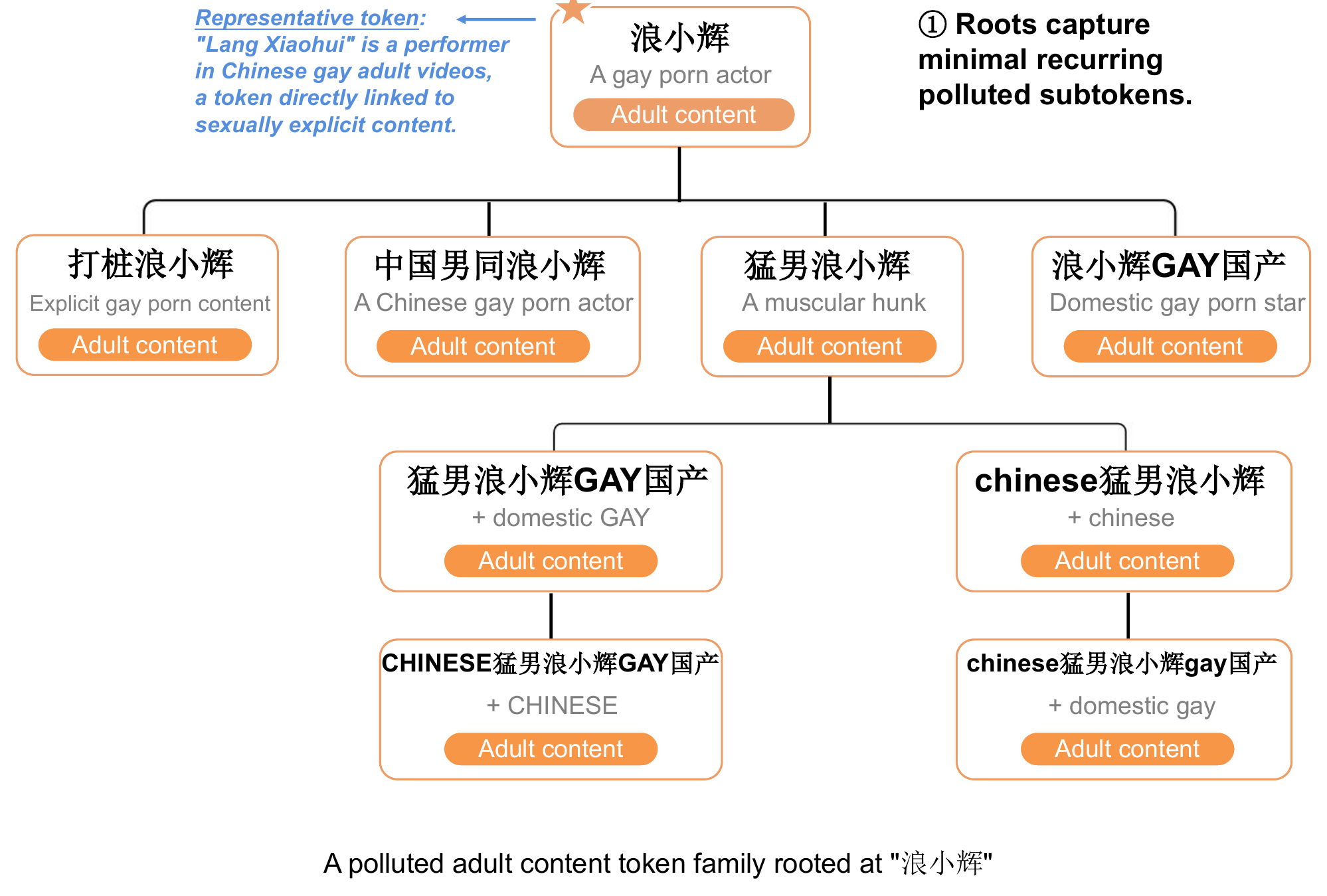}
  \caption{Token tree rooted at ``浪小辉''. The root captures a recurring Adult Content entity, and descendants show longer surface variants.}
  \label{fig:token-tree-xiaohui}
\end{figure*}

\autoref{fig:token-tree-xiangjiao} and \autoref{fig:token-tree-xiaohui} illustrate the polluted-root pattern. In \autoref{fig:token-tree-xiangjiao}, ``大道香蕉'' functions as a recurring adult-site keyword, and the descendants add modifiers such as duration, language, online viewing, and video type. In \autoref{fig:token-tree-xiaohui}, ``浪小辉'' refers to an adult-content entity, and the descendants add identity, regional, and content-type variants. In both cases, the root itself is already polluted, so the tree acts as a compact summary of many longer surface forms. This makes the dataset easier to review and helps future cleaning systems generalize beyond exact-string matching.

\begin{figure*}[t]
  \centering
  \includegraphics[width=\textwidth]{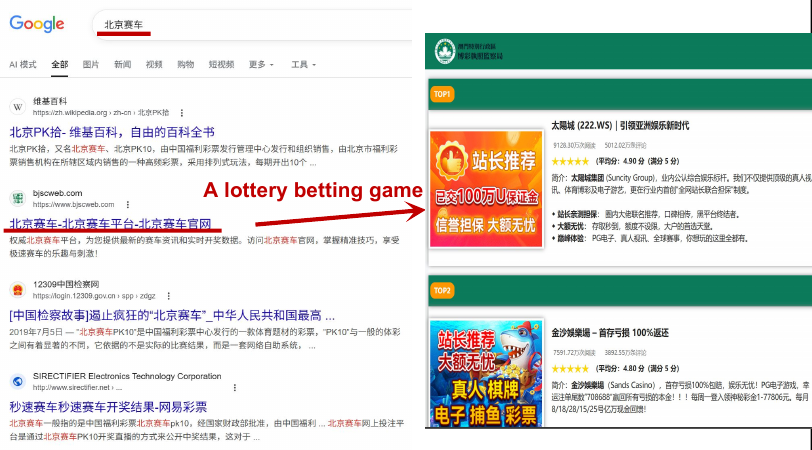}
  \caption{Search evidence for ``北京赛车'', showing web context to identify the token family as Online Gambling.}
  \label{fig:search-evidence-saiche}
\end{figure*}

\begin{figure*}[t]
  \centering
  \includegraphics[width=\textwidth]{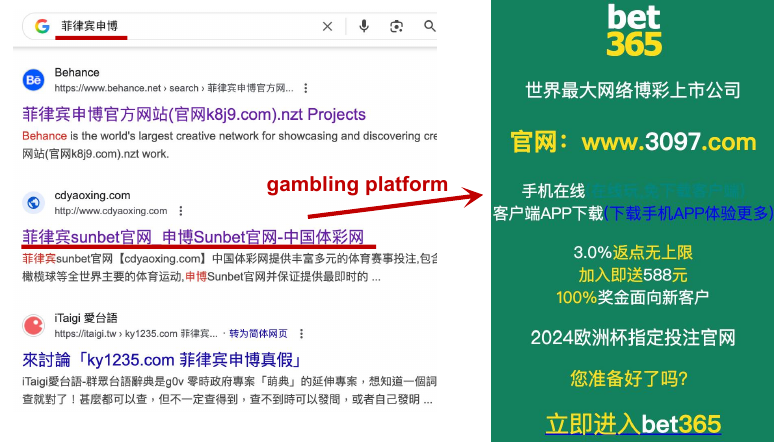}
  \caption{Search evidence for ``菲律宾申博'', showing web context to identify the token family as Gambling.}
  \label{fig:search-evidence-shenbo}
\end{figure*}

\begin{figure*}[t]
  \centering
  \includegraphics[width=\textwidth]{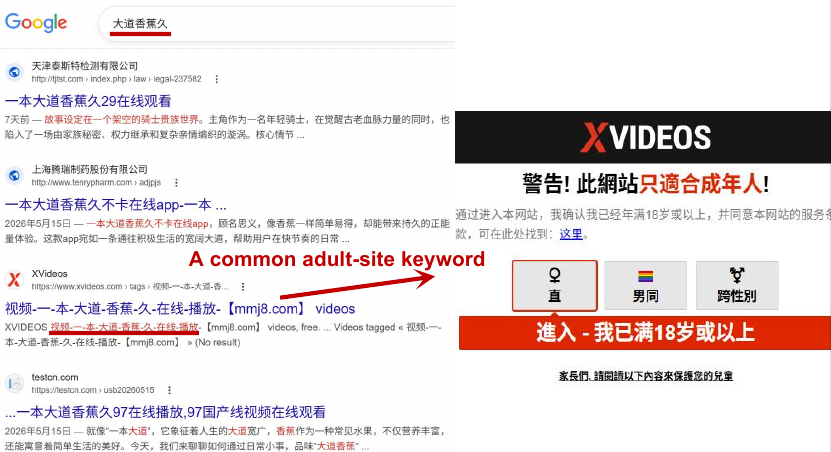}
  \caption{Search evidence for ``大道香蕉'', showing web context to identify the token family as Adult Content.}
  \label{fig:search-evidence-xiangjiao}
\end{figure*}

\begin{figure*}[t]
  \centering
  \includegraphics[width=\textwidth]{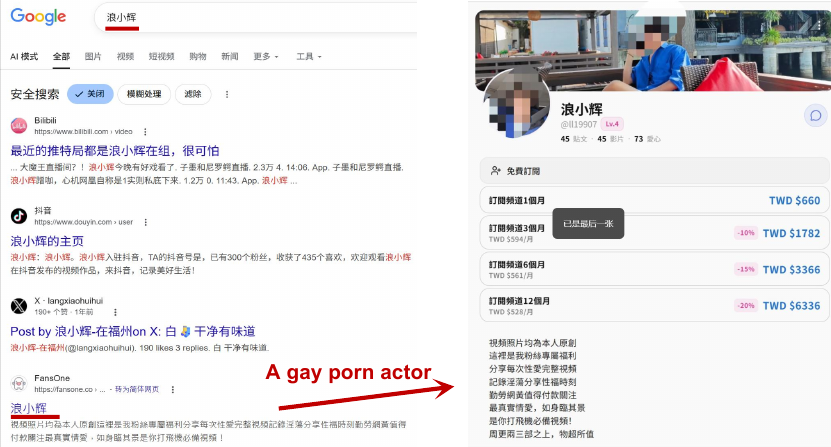}
  \caption{Search evidence for ``浪小辉'', showing web context used to identify the token family as Adult Content.}
  \label{fig:search-evidence-xiaohui}
\end{figure*}

The corresponding search-evidence figures provide the contextual basis for these labels. \autoref{fig:search-evidence-saiche} and \autoref{fig:search-evidence-shenbo} show that the gambling labels are supported by search results and webpages connected to lottery betting, gambling platforms, and betting-site promotion. \autoref{fig:search-evidence-xiangjiao} and \autoref{fig:search-evidence-xiaohui} show that the adult-content labels are supported by adult-site results and explicit-content contexts. These examples illustrate a recurring difficulty in Chinese web-pollution auditing: many polluted tokens are short, euphemistic, or overloaded with benign meanings. Web context is therefore necessary for auditable category assignment, while the tree structure makes the resulting decisions traceable across related token variants.

\end{CJK}
\end{document}

%% file: preamble.tex
\usepackage{latexsym}
\usepackage{adjustbox}
\usepackage{xspace}
\usepackage{todonotes}
\usepackage{xcolor}
\usepackage{comment}
\usepackage{tikz}
\usetikzlibrary{arrows.meta,positioning,calc}
\usepackage{pgfplots}
\usepackage{pgfplotstable}
\usepackage{makecell}
\usepackage{adjustbox}
\usepackage{pifont}
\usepackage{float}
\usepackage{supertabular}
\usepackage{multirow}
\usepackage{longtable}
\usepackage{graphicx}
\usepackage[edges]{forest}
\definecolor{hidden-draw}{RGB}{0,0,0}
\usepackage{wrapfig}
\usepackage{CJKutf8}
\usepackage{CJK}

\usepackage{tcolorbox}
\tcbset{
    mybox/.style={
        fontupper=\linespread{.8}\selectfont,
        sharp corners,
        top=0pt, 
        bottom=0pt 
    }
}

\usepackage{amsmath}
\usepackage{amssymb}

\usepackage{booktabs}
\usepackage{threeparttable}
\usepackage{tabularx}

\usepackage{soul}

\usepackage{bm} 

\usepackage{algorithm}
\usepackage{algorithmic}

\usepackage{setspace}

\definecolor{mygreen}{RGB}{11,141,10}
\definecolor{myred}{RGB}{223,68,52}
\definecolor{myblue}{RGB}{70,130,180}
\definecolor{mydeepblue}{RGB}{65,105,225}
\definecolor{myviolet}{RGB}{97,0,138}
\definecolor{myburgundy}{RGB}{110,10,30}
\definecolor{myblue2}{RGB}{0,105,148}
\definecolor{iceblue}{RGB}{173, 216, 230}
\definecolor{puregreen}{RGB}{0, 218, 0}
\definecolor{graygreen}{RGB}{74,113,106}

\definecolor{wingreen}{rgb}{0,0.45,0.24}
\definecolor{losered}{rgb}{1.0,0.1,0.24}

\definecolor{lightcoral}{rgb}{0.97, 0.36, 0.46}
\definecolor{lightyellow}{rgb}{0.98, 0.7, 0}
\definecolor{harvestgold}{rgb}{0.85, 0.57, 0.0}
\definecolor{brightlavender}{rgb}{0.75, 0.58, 0.89}
\definecolor{capri}{rgb}{0.0, 0.75, 1.0}
\definecolor{carminepink}{rgb}{0.92, 0.3, 0.26}
\definecolor{celadon}{rgb}{0.67, 0.88, 0.69}
\definecolor{darkpastelgreen}{rgb}{0.01, 0.75, 0.24}

\definecolor{grayhighlight}{RGB}{250,250,227}

\definecolor{target}{HTML}{F47983}
\definecolor{control}{HTML}{3E87CD}
\definecolor{credibility}{HTML}{B98AC9}
\definecolor{logical}{HTML}{93C572}
\definecolor{emotional}{HTML}{F9EAC3}

\newenvironment{packeditemize}{
\begin{list}{$\bullet$}{
\setlength{\labelwidth}{8pt}
\setlength{\itemsep}{0pt}
\setlength{\leftmargin}{\labelwidth}
\addtolength{\leftmargin}{\labelsep}
\setlength{\parindent}{0pt}
\setlength{\listparindent}{\parindent}
\setlength{\parsep}{0pt}
\setlength{\topsep}{3pt}}}{\end{list}}

\usepackage{pifont}

\newcommand{\diffdown}[1]{\raisebox{0.5pt}{\fontsize{6}{5.5}\selectfont{\textcolor{wingreen}{\textbf{$\blacktriangledown${$ #1$}}}}}}

\newcommand{\usepalatino}[1]{{\fontfamily{ppl}\selectfont #1}}

\newcommand{\warning}{\raisebox{2pt}{\fontencoding{U}\fontfamily{futs}\selectfont\char 49\relax}}

\newcommand{\myline}{\par
  \kern0pt 
  \hrule height 0.6pt
  \kern3pt 
}

\newcommand{\mylinenoskip}{\par
  \kern3pt 
  \hrule height 0.6pt
  \kern3pt 
}

\usepackage{amsthm}

\usepackage{url}
\usepackage{booktabs}
\usepackage{siunitx}

%% file: custom.bib
@article{raffel2020exploring,
  title={Exploring the limits of transfer learning with a unified text-to-text transformer},
  author={Raffel, Colin and Shazeer, Noam and Roberts, Adam and Lee, Katherine and Narang, Sharan and Matena, Michael and Zhou, Yanqi and Li, Wei and Liu, Peter J},
  journal={Journal of machine learning research},
  volume={21},
  number={140},
  pages={1--67},
  year={2020}
}

@article{penedo2023refinedweb,
  title={The refinedweb dataset for falcon llm: outperforming curated corpora with web data, and web data only},
  author={Penedo, Guilherme and Malartic, Quentin and Hesslow, Daniel and Cojocaru, Ruxandra and Cappelli, Alessandro and Alobeidli, Hamza and Pannier, Baptiste and Almazrouei, Ebtesam and Launay, Julien},
  journal={arXiv preprint arXiv:2306.01116},
  year={2023}
}

@article{weber2024redpajama,
  title={Redpajama: an open dataset for training large language models},
  author={Weber, Maurice and Fu, Daniel Y and Anthony, Quentin and Oren, Yonatan and Adams, Shane and Alexandrov, Anton and Lyu, Xiaozhong and Nguyen, Huu and Yao, Xiaozhe and Adams, Virginia and others},
  journal={Advances in neural information processing systems},
  volume={37},
  pages={116462--116492},
  year={2024}
}

@article{penedo2024fineweb,
  title={The fineweb datasets: Decanting the web for the finest text data at scale},
  author={Penedo, Guilherme and Kydl{\'\i}{\v{c}}ek, Hynek and Lozhkov, Anton and Mitchell, Margaret and Raffel, Colin and Von Werra, Leandro and Wolf, Thomas and others},
  journal={Advances in Neural Information Processing Systems},
  volume={37},
  pages={30811--30849},
  year={2024}
}

@inproceedings{xu-etal-2025-infini,
  title={INFINI-GRAM MINI: Exact n-gram Search at the Internet Scale with FM-Index},
  author={Xu, Hao and Liu, Jiacheng and Choi, Yejin and Smith, Noah A and Hajishirzi, Hannaneh},
  booktitle={Proceedings of the 2025 Conference on Empirical Methods in Natural Language Processing},
  pages={24955--24980},
  year={2025}
}

@inproceedings{soldaini2024dolma,
  title={Dolma: An open corpus of three trillion tokens for language model pretraining research},
  author={Soldaini, Luca and Kinney, Rodney and Bhagia, Akshita and Schwenk, Dustin and Atkinson, David and Authur, Russell and Bogin, Ben and Chandu, Khyathi and Dumas, Jennifer and Elazar, Yanai and others},
  booktitle={Proceedings of the 62nd Annual Meeting of the Association for Computational Linguistics (Volume 1: Long Papers)},
  pages={15725--15788},
  year={2024}
}

@misc{commoncrawl,
  author = {{Common Crawl}},
  title  = {Common Crawl},
  year   = {2011},
  howpublished = {\url{https://commoncrawl.org/}}
}

@article{he2023wanjuan,
  title={Wanjuan: A comprehensive multimodal dataset for advancing english and chinese large models},
  author={He, Conghui and Jin, Zhenjiang and Xu, Chao and Qiu, Jiantao and Wang, Bin and Li, Wei and Yan, Hang and Wang, Jiaqi and Lin, Dahua},
  journal={arXiv preprint arXiv:2308.10755},
  year={2023}
}

@inproceedings{abadji2022oscar,
  title={Towards a cleaner document-oriented multilingual crawled corpus},
  author={Abadji, Julien and Suarez, Pedro Ortiz and Romary, Laurent and Sagot, Beno{\^\i}t},
  booktitle={Proceedings of the Thirteenth Language Resources and Evaluation Conference},
  pages={4344--4355},
  year={2022}
}

@misc{oscar2301hpc,
  author = {{OSCAR Project}},
  title = {{OSCAR-2301-HPC}},
  year = {2023},
  howpublished = {\url{https://huggingface.co/datasets/oscar-corpus/oscar-2301-hpc}},
  note = {Hugging Face dataset card}
}

@inproceedings{xue2021mt5,
  title={mT5: A massively multilingual pre-trained text-to-text transformer},
  author={Xue, Linting and Constant, Noah and Roberts, Adam and Kale, Mihir and Al-Rfou, Rami and Siddhant, Aditya and Barua, Aditya and Raffel, Colin},
  booktitle={Proceedings of the 2021 conference of the North American chapter of the association for computational linguistics: Human language technologies},
  pages={483--498},
  year={2021}
}

@misc{allenai_c4,
  author = {{Allen Institute for AI}},
  title = {{C4}: Colossal Clean Crawled Corpus},
  year = {2024},
  howpublished = {\url{https://huggingface.co/datasets/allenai/c4}},
  note = {Hugging Face dataset card}
}

@article{oepen2025hplt3,
  title={HPLT 3.0: Very Large-Scale Multilingual Resources for LLM and MT. Mono-and Bi-lingual Data, Multilingual Evaluation, and Pre-Trained Models},
  author={Oepen, Stephan and Arefev, Nikolay and Aulamo, Mikko and Ba{\~n}{\'o}n, Marta and Buljan, Maja and Burchell, Laurie and Charpentier, Lucas and Chen, Pinzhen and Fedorova, Mariya and de Gibert, Ona and others},
  journal={arXiv preprint arXiv:2511.01066},
  year={2025}
}

@misc{hplt3_catalog,
  author = {{High Performance Language Technologies Project}},
  title = {{HPLT} Monolingual Datasets 3.0},
  year = {2025},
  howpublished = {\url{https://hplt-project.org/datasets/v3.0}},
  note = {Dataset catalogue}
}

@inproceedings{nguy2024culturax,
    title = "{C}ultura{X}: A Cleaned, Enormous, and Multilingual Dataset for Large Language Models in 167 Languages",
    author = "Nguyen, Thuat  and
      Nguyen, Chien Van  and
      Lai, Viet Dac  and
      Man, Hieu  and
      Ngo, Nghia Trung  and
      Dernoncourt, Franck  and
      Rossi, Ryan A.  and
      Nguyen, Thien Huu",
    editor = "Calzolari, Nicoletta  and
      Kan, Min-Yen  and
      Hoste, Veronique  and
      Lenci, Alessandro  and
      Sakti, Sakriani  and
      Xue, Nianwen",
    booktitle = "Proceedings of the 2024 Joint International Conference on Computational Linguistics, Language Resources and Evaluation (LREC-COLING 2024)",
    month = may,
    year = "2024",
    address = "Torino, Italia",
    publisher = "ELRA and ICCL",
    url = "https://aclanthology.org/2024.lrec-main.377/",
    pages = "4226--4237"
}

@article{chen2023chinesewebtext,
  title={Chinesewebtext: Large-scale high-quality Chinese web text extracted with effective evaluation model},
  author={Chen, Jianghao and Jian, Pu and Xi, Tengxiao and Yi, Dongyi and Du, Qianlong and Ding, Chenglin and Zhu, Guibo and Zong, Chengqing and Wang, Jinqiao and Zhang, Jiajun},
  journal={arXiv preprint arXiv:2311.01149},
  year={2023}
}

@article{laurencon2022roots,
  title={The bigscience roots corpus: A 1.6 tb composite multilingual dataset},
  author={Lauren{\c{c}}on, Hugo and Saulnier, Lucile and Wang, Thomas and Akiki, Christopher and Villanova del Moral, Albert and Le Scao, Teven and Von Werra, Leandro and Mou, Chenghao and Gonz{\'a}lez Ponferrada, Eduardo and Nguyen, Huu and others},
  journal={Advances in Neural Information Processing Systems},
  volume={35},
  pages={31809--31826},
  year={2022}
}

@inproceedings{rootssearch2023,
  title={The ROOTS search tool: Data transparency for LLMs},
  author={Piktus, Aleksandra and Akiki, Christopher and Villegas, Paulo and Lauren{\c{c}}on, Hugo and Dupont, G{\'e}rard and Luccioni, Sasha and Jernite, Yacine and Rogers, Anna},
  booktitle={Proceedings of the 61st Annual Meeting of the Association for Computational Linguistics (Volume 3: System Demonstrations)},
  pages={304--314},
  year={2023}
}

@article{bigscience2022bloom,
  title={Bloom: A 176b-parameter open-access multilingual language model},
  author={Workshop, BigScience and Scao, Teven Le and Fan, Angela and Akiki, Christopher and Pavlick, Ellie and Ili{\'c}, Suzana and Hesslow, Daniel and Castagn{\'e}, Roman and Luccioni, Alexandra Sasha and Yvon, Fran{\c{c}}ois and others},
  journal={arXiv preprint arXiv:2211.05100},
  year={2022}
}

@article{du2024chinesetinyllm,
  title={Chinese tiny llm: Pretraining a chinese-centric large language model},
  author={Du, Xinrun and Yu, Zhouliang and Gao, Songyang and Pan, Ding and Cheng, Yuyang and Ma, Ziyang and Yuan, Ruibin and Qu, Xingwei and Liu, Jiaheng and Zheng, Tianyu and others},
  journal={arXiv preprint arXiv:2404.04167},
  year={2024}
}

@misc{mapcc_dataset,
      title={Chinese Tiny LLM: Pretraining a Chinese-Centric Large Language Model}, 
      author={Xinrun Du and Zhouliang Yu and Songyang Gao and Ding Pan and Yuyang Cheng and Ziyang Ma and Ruibin Yuan and Xingwei Qu and Jiaheng Liu and Tianyu Zheng and Xinchen Luo and Guorui Zhou and Wenhu Chen and Ge Zhang},
      year={2024},
      eprint={2404.04167},
      archivePrefix={arXiv},
      primaryClass={cs.CL},
      url={https://arxiv.org/abs/2404.04167}, 
}

@article{wei2023skywork,
  title={Skywork: A more open bilingual foundation model},
  author={Wei, Tianwen and Zhao, Liang and Zhang, Lichang and Zhu, Bo and Wang, Lijie and Yang, Haihua and Li, Biye and Cheng, Cheng and L{\"u}, Weiwei and Hu, Rui and others},
  journal={arXiv preprint arXiv:2310.19341},
  year={2023}
}

@article{wang2024cci30hq,
  title={CCI3. 0-HQ: a large-scale Chinese dataset of high quality designed for pre-training large language models},
  author={Wang, Liangdong and Zhang, Bo-Wen and Wu, Chengwei and Zhao, Hanyu and Shi, Xiaofeng and Gu, Shuhao and Li, Jijie and Ma, Quanyue and Pan, TengFei and Liu, Guang},
  journal={arXiv preprint arXiv:2410.18505},
  year={2024}
}

@article{yuan2021wudaocorpora,
  title={Wudaocorpora: A super large-scale chinese corpora for pre-training language models},
  author={Yuan, Sha and Zhao, Hanyu and Du, Zhengxiao and Ding, Ming and Liu, Xiao and Cen, Yukuo and Zou, Xu and Yang, Zhilin and Tang, Jie},
  journal={Ai Open},
  volume={2},
  pages={65--68},
  year={2021},
  publisher={Elsevier}
}

@misc{wudaocorpusbase,
  author = {{PaddleNLP Contributors}},
  title = {{WuDaoCorpus2.0} Base Corpus},
  year = {2025},
  howpublished = {\url{https://paddlenlp.readthedocs.io/en/latest/llm/tools/preprocess/docs/WuDaoCorpusBase.html}},
  note = {Documentation page}
}

@article{gebru2021datasheets,
  title={Datasheets for datasets},
  author={Gebru, Timnit and Morgenstern, Jamie and Vecchione, Briana and Vaughan, Jennifer Wortman and Wallach, Hanna and Iii, Hal Daum{\'e} and Crawford, Kate},
  journal={Communications of the ACM},
  volume={64},
  number={12},
  pages={86--92},
  year={2021},
  publisher={ACM New York, NY, USA}
}

@inproceedings{dodge2021documentingc4,
  title={Documenting large webtext corpora: A case study on the colossal clean crawled corpus},
  author={Dodge, Jesse and Sap, Maarten and Marasovi{\'c}, Ana and Agnew, William and Ilharco, Gabriel and Groeneveld, Dirk and Mitchell, Margaret and Gardner, Matt},
  booktitle={Proceedings of the 2021 conference on empirical methods in natural language processing},
  pages={1286--1305},
  year={2021}
}

@article{kreutzer2022quality,
  title={Quality at a glance: An audit of web-crawled multilingual datasets},
  author={Kreutzer, Julia and Caswell, Isaac and Wang, Lisa and Wahab, Ahsan and Van Esch, Daan and Ulzii-Orshikh, Nasanbayar and Tapo, Allahsera and Subramani, Nishant and Sokolov, Artem and Sikasote, Claytone and others},
  journal={Transactions of the Association for Computational Linguistics},
  volume={10},
  pages={50--72},
  year={2022},
  publisher={MIT Press One Rogers Street, Cambridge, MA 02142-1209, USA journals-info~…}
}

@inproceedings{lee2022deduplicating,
  title={Deduplicating training data makes language models better},
  author={Lee, Katherine and Ippolito, Daphne and Nystrom, Andrew and Zhang, Chiyuan and Eck, Douglas and Callison-Burch, Chris and Carlini, Nicholas},
  booktitle={Proceedings of the 60th Annual Meeting of the Association for Computational Linguistics (Volume 1: Long Papers)},
  pages={8424--8445},
  year={2022}
}

@article{abbas2023semdedup,
  title={Semdedup: Data-efficient learning at web-scale through semantic deduplication},
  author={Abbas, Amro and Tirumala, Kushal and Simig, D{\'a}niel and Ganguli, Surya and Morcos, Ari S},
  journal={arXiv preprint arXiv:2303.09540},
  year={2023}
}

@inproceedings{zhang2025speculating,
  title={Speculating LLMs’ Chinese Training Data Pollution from Their Tokens},
  author={Zhang, Qingjie and Wang, Di and Qian, Haoting and Yan, Liu and Zhang, Tianwei and Xu, Ke and Li, Qi and Huang, Minlie and Li, Hewu and Qiu, Han},
  booktitle={Proceedings of the 2025 Conference on Empirical Methods in Natural Language Processing},
  pages={26124--26144},
  year={2025}
}

@inproceedings{hagar2025practical,
  title={Practical Datasets for Analyzing LLM Corpora Derived from Common Crawl},
  author={Hagar, Nick and Bandy, Jack},
  booktitle={Proceedings of the International AAAI Conference on Web and Social Media},
  volume={19},
  pages={2454--2464},
  year={2025}
}

@inproceedings{li2025ticlm,
  title={TiC-LM: A web-scale benchmark for time-continual LLM pretraining},
  author={Li, Jeffrey and Armandpour, Mohammadreza and Mirzadeh, Seyed Iman and Mehta, Sachin and Shankar, Vaishaal and Vemulapalli, Raviteja and Bengio, Samy and Tuzel, Oncel and Farajtabar, Mehrdad and Pouransari, Hadi and others},
  booktitle={Proceedings of the 63rd Annual Meeting of the Association for Computational Linguistics (Volume 1: Long Papers)},
  pages={32231--32273},
  year={2025}
}

@inproceedings{wang2018glue,
  title={GLUE: A multi-task benchmark and analysis platform for natural language understanding},
  author={Wang, Alex and Singh, Amanpreet and Michael, Julian and Hill, Felix and Levy, Omer and Bowman, Samuel},
  booktitle={Proceedings of the 2018 EMNLP workshop BlackboxNLP: Analyzing and interpreting neural networks for NLP},
  pages={353--355},
  year={2018}
}

@inproceedings{longpre2023flan,
  title={The flan collection: Designing data and methods for effective instruction tuning},
  author={Longpre, Shayne and Hou, Le and Vu, Tu and Webson, Albert and Chung, Hyung Won and Tay, Yi and Zhou, Denny and Le, Quoc V and Zoph, Barret and Wei, Jason and others},
  booktitle={International conference on machine learning},
  pages={22631--22648},
  year={2023},
  organization={PMLR}
}

@article{brown2020language,
  title={Language models are few-shot learners},
  author={Brown, Tom and Mann, Benjamin and Ryder, Nick and Subbiah, Melanie and Kaplan, Jared D and Dhariwal, Prafulla and Neelakantan, Arvind and Shyam, Pranav and Sastry, Girish and Askell, Amanda and others},
  journal={Advances in neural information processing systems},
  volume={33},
  pages={1877--1901},
  year={2020}
}

@inproceedings{roziewski2016languagecrawl,
  title={Languagecrawl: A generic tool for building language models upon common-crawl},
  author={Roziewski, Szymon and Stokowiec, Wojciech},
  booktitle={Proceedings of the Tenth International Conference on Language Resources and Evaluation (LREC'16)},
  pages={2789--2793},
  year={2016}
}

@misc{yang2024gpt4oChineseTokenPollution,
  author = {Yang, Zeyi},
  title = {{GPT-4o}'s {Chinese} Token-Training Data Is Polluted by Spam and Porn Websites},
  year = {2024},
  month = may,
  howpublished = {{MIT Technology Review}},
  url = {https://www.technologyreview.com/2024/05/17/1092649/gpt-4o-chinese-token-polluted/},
  note = {Published May 17, 2024}
}

@misc{openaiCommunity2026codexChineseGambling,
  author = {{OpenAI Developer Community}},
  title = {Chinese Gambling Characters in {Codex CLI} Message and Code Output?},
  year = {2026},
  month = jan,
  howpublished = {{OpenAI Developer Community}},
  url = {https://community.openai.com/t/chinese-gambling-characters-in-codex-cli-message-and-code-output/1372678},
  note = {Forum thread, first posted January 27, 2026}
}

@inproceedings{sennrich2016neural,
  title={Neural machine translation of rare words with subword units},
  author={Sennrich, Rico and Haddow, Barry and Birch, Alexandra},
  booktitle={Proceedings of the 54th annual meeting of the association for computational linguistics (volume 1: long papers)},
  pages={1715--1725},
  year={2016}
}

@article{karydis2013marinewater,
  title={Marine water quality monitoring: A review},
  author={Karydis, Michael and Kitsiou, Dimitra},
  journal={Marine pollution bulletin},
  volume={77},
  number={1-2},
  pages={23--36},
  year={2013},
  publisher={Elsevier}
}

@article{jiang2021neologisms,
  title={Neologisms are epidemic: Modeling the life cycle of neologisms in China 2008-2016},
  author={Jiang, Menghan and Shen, Xiang Ying and Ahrens, Kathleen and Huang, Chu-Ren},
  journal={PloS one},
  volume={16},
  number={2},
  pages={e0245984},
  year={2021},
  publisher={Public Library of Science San Francisco, CA USA}
}

@article{macavaney2019hate,
  title={Hate speech detection: Challenges and solutions},
  author={MacAvaney, Sean and Yao, Hao-Ren and Yang, Eugene and Russell, Katina and Goharian, Nazli and Frieder, Ophir},
  journal={PloS one},
  volume={14},
  number={8},
  pages={e0221152},
  year={2019},
  publisher={Public Library of Science San Francisco, CA USA}
}
